\documentclass[10pt,twocolumn,letterpaper]{article}

\usepackage{cvpr}              

\usepackage[accsupp]{axessibility}  

\usepackage{graphicx}
\usepackage{amsmath}
\usepackage{amssymb}
\usepackage{booktabs}

\usepackage{graphicx}
\usepackage{amsmath,amsfonts,bm}
\usepackage{booktabs}
\usepackage{algorithm}
\usepackage{algorithmic}

\usepackage{amsmath,amsfonts,bm}

\def\eqref#1{equation~\ref{#1}}

\def\1{\bm{1}}

\DeclareMathAlphabet{\mathsfit}{\encodingdefault}{\sfdefault}{m}{sl}
\SetMathAlphabet{\mathsfit}{bold}{\encodingdefault}{\sfdefault}{bx}{n}

\DeclareMathOperator*{\argmax}{arg\,max}

\usepackage{bbding}
\usepackage{multirow}

\usepackage{booktabs}
\usepackage{array}
\usepackage{soul}
\usepackage[flushleft]{threeparttable}
\def\bs{\bm}

\usepackage[pagebackref,breaklinks,colorlinks]{hyperref}

\usepackage[capitalize]{cleveref}
\crefname{section}{Sec.}{Secs.}
\Crefname{section}{Section}{Sections}
\Crefname{table}{Table}{Tables}
\crefname{table}{Tab.}{Tabs.}

\def\cvprPaperID{2354} 
\def\confName{CVPR}
\def\confYear{2023}

\begin{document}

\title{Towards Bridging the Performance Gaps of Joint Energy-based Models}

\author{Xiulong Yang, Qing Su, and Shihao Ji\\
Georgia State University\\
{\tt\small \{xyang22,qsu3,sji\}@gsu.edu}
}
\maketitle

\begin{abstract}
  Can we train a hybrid discriminative-generative model with a single network? This question has recently been answered in the affirmative, introducing the field of Joint Energy-based Model (JEM)~\cite{jem,jempp}, which achieves high classification accuracy and image generation quality simultaneously. Despite recent advances, there remain two performance gaps: the accuracy gap to the standard softmax classifier, and the generation quality gap to state-of-the-art generative models. In this paper,  we introduce a variety of training techniques to bridge the accuracy gap and the generation quality gap of JEM. 1) We incorporate a recently proposed sharpness-aware minimization (SAM) framework to train JEM, which promotes the energy landscape smoothness and the generalization of JEM. 2) We exclude data augmentation from the maximum likelihood estimate pipeline of JEM, and mitigate the negative impact of data augmentation to image generation quality. Extensive experiments on multiple datasets demonstrate our SADA-JEM achieves state-of-the-art performances and outperforms JEM in image classification, image generation, calibration, out-of-distribution detection and adversarial robustness by a notable margin. Our code is available at \url{https://github.com/sndnyang/SADAJEM}.
\end{abstract}


\section{Introduction}

Deep neural networks (DNNs) have achieved state-of-the-art performances in a wide range of learning tasks, including image classification, image generation, object detection, and language understanding~\cite{Krizhevsky2012,resnet16}. Among them, energy-based models (EBMs) have seen a flurry of interest recently, partially inspired by the impressive results of IGEBM~\cite{du2019implicit} and JEM~\cite{jem}, which exhibit the capability of training generative models within a discriminative framework. Specifically, JEM~\cite{jem} reinterprets the standard softmax classifier as an EBM and achieves impressive performances in image classification and generation simultaneously. Furthermore, these EBMs enjoy improved performance on out-of-distribution detection, calibration, and adversarial robustness. The follow-up works (e.g.,~\cite{jempp,nomcmc}) further improve the training in terms of speed, stability and accuracy.

\begin{figure*}
  \centering
  \begin{subfigure}{0.23\linewidth}
    \includegraphics[width=1\columnwidth]{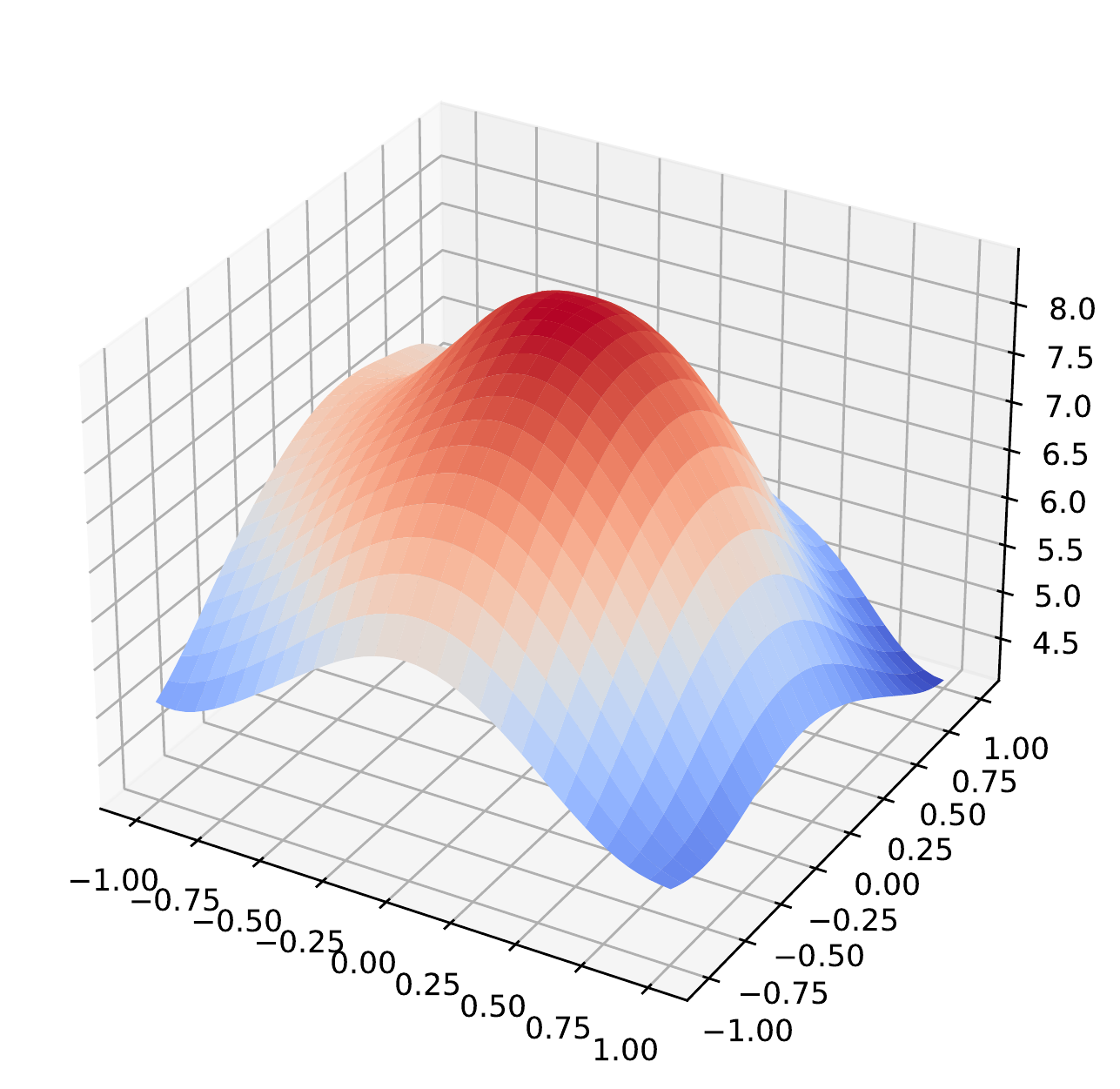}
    \caption{Softmax Classifier}
    \label{figure:e_landscape_cls}
  \end{subfigure}
  \hfill
  \begin{subfigure}{0.23\linewidth}
    \includegraphics[width=1\columnwidth]{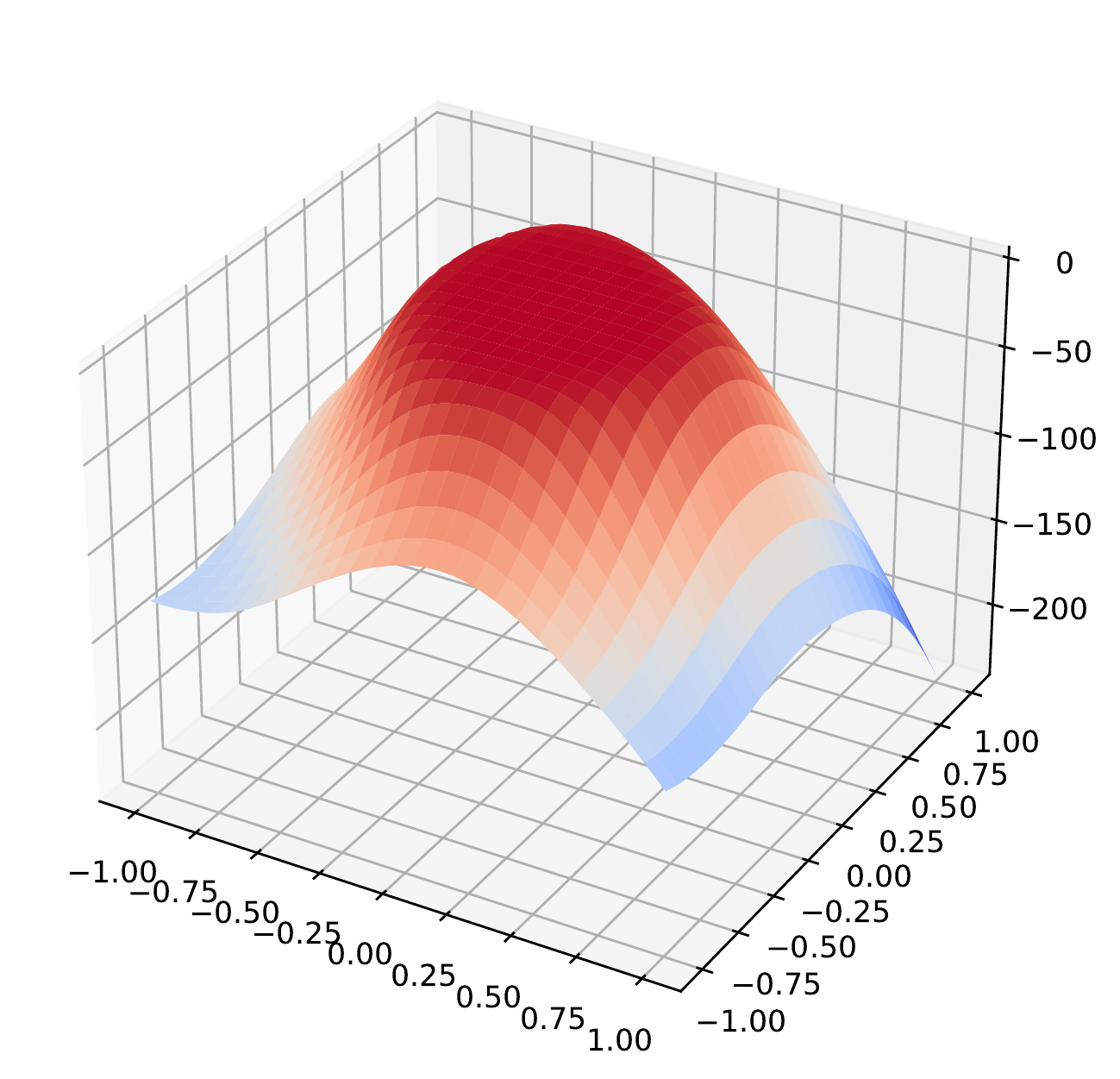}
    \caption{JEM}
    \label{figure:e_landscape_official_jem}
  \end{subfigure}
  \hfill
  \begin{subfigure}{0.23\linewidth}
    \includegraphics[width=1\columnwidth]{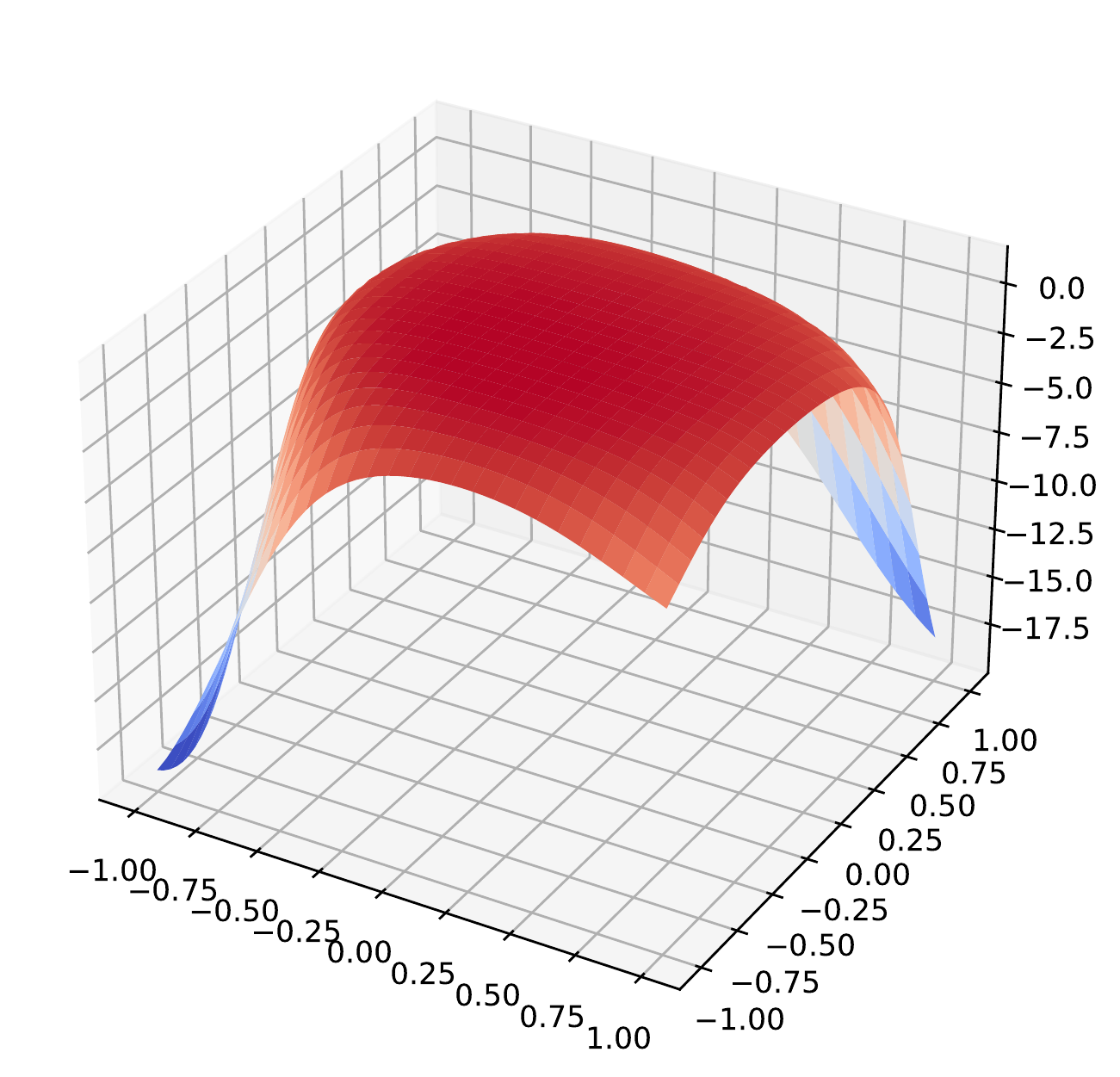}
    \caption{JEM+SAM}
    \label{figure:e_landscape_jemsam}
  \end{subfigure}
  \hfill
  \begin{subfigure}{0.23\linewidth}
    \includegraphics[width=1\columnwidth]{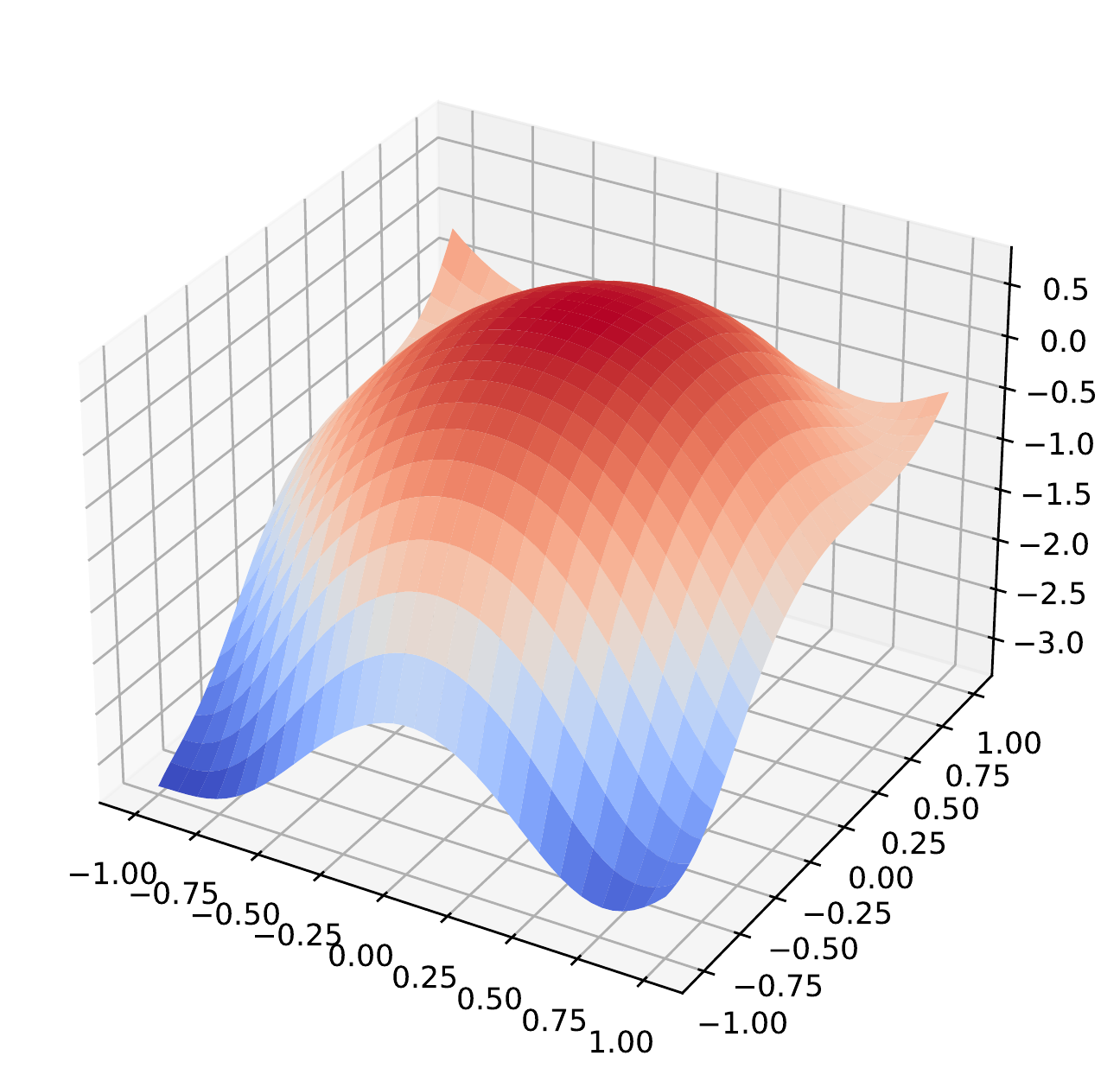}
    \caption{SADA-JEM}
    \label{figure:e_landscape_sajem10}
  \end{subfigure}

    \caption{Visualizing the energy landscapes~\cite{losslandscape} of different models trained on CIFAR10. Note the dramatic scale differences of the y-axes, indicating SADA-JEM identifies the smoothest local optimum among all the methods considered.}
    \label{figure:energy_landscape}
\end{figure*}

Despite the recent advances and the appealing property of training a single network for hybrid modeling, training JEM is still challenging on complex high-dimensional data since it requires an expensive MCMC sampling. Furthermore, models produced by JEM still have an accuracy gap to the standard softmax classifier and a generation quality gap to the GAN-based approaches.

In this paper, we introduce a few simple yet effective training techniques to bridge the accuracy gap and generation quality gap of JEM. Our hypothesis is that both performance gaps are the symptoms of lack of generalization of JEM trained models. We therefore analyze the trained models under the lens of loss geometry. Figure~\ref{figure:energy_landscape} visualizes the energy landscapes of different models by the technique introduced in~\cite{losslandscape}. Since different models are trained with different loss functions, visualizing their loss functions is meaningless for the purpose of comparison. Therefore, the LSE energy functions (i.e., Eq.~\ref{eq:jem_ex}) of different models are visualized. Comparing Figure~\ref{figure:energy_landscape}(a) and (b), we find that JEM converges to extremely sharp local maxima of the energy landscape as manifested by the significantly large y-axis scale. By incorporating the recently proposed sharpness-aware minimization (SAM)~\cite{sam2021} to JEM, the energy landscape of trained model (JEM+SAM) becomes much smoother as shown in Figure~\ref{figure:energy_landscape}(c). This also substantially improves the image classification accuracy and generation quality. To further improve the energy landscape smoothness, we exclude data augmentation from the maximum likelihood estimate pipeline of JEM, and visualize the energy landscape of SADA-JEM in  Figure~\ref{figure:energy_landscape}(d), which achieves the smoothest landscape among all the models considered. This further improves image generation quality dramatically while retaining or sometimes improving classification accuracy. Since our method improves the performance of JEM primarily in the framework of sharpness-aware optimization, we refer it as SADA-JEM, a Sharpness-Aware Joint Energy-based Model with single branched Data Augmentation.

Our main contributions are summarized as follows:
\begin{enumerate}
\item We investigate the energy landscapes of different models and find that JEM leads to the sharpest one, which potentially undermines the generalization of trained models.

\item We incorporate the sharpness-aware minimization (SAM) framework to JEM to promote the energy landscape smoothness, and thus model generalization.

\item We recognize the negative impact of data augmentation in the training pipeline of JEM, and introduce two data loaders for image classification and image generation separately, which improves image generation quality significantly.

\item Extensive experiments on multiple datasets show that SADA-JEM achieves the state-of-the-art discriminative and generative performances, while outperforming JEM in calibration, out-of-distribution detection and adversarial robustness by a notable margin.
\end{enumerate}

\section{Related Work}
\paragraph{Energy-Based Models} (EBMs)~\cite{lecun2006tutorial} stem from the observation that any probability density function $p_{\bs{\theta}}(\bs{x})$ can be expressed via a Boltzmann distribution as
\begin{equation}\label{eq:ebm_define}
  p_{\bs{\theta}}(\bs{x})=\frac{\exp \left(-E_{\bs{\theta}}(\bs{x})\right)}{Z(\bs{\theta})},
\end{equation}
where $E_{\bs{\theta}}(\bs{x})$ is an energy function that maps input $\bs{x}\in\mathcal{X}$ to a scalar, and $Z(\bs{\theta}) = \int_{\bs{x}} \exp \left(-E_{\bs{\theta}}(\bs{x})\right)$ is the normalizing constant w.r.t. $\bs{x}$ (also known as the partition function). Ideally, an energy function should assign low energy values to the samples drawn from data distribution, and high values otherwise.

The key challenge of EBM training is to estimate the intractable partition function $Z(\bs{\theta})$, and thus the maximum likelihood estimate of parameters $\bs{\theta}$ is not straightforward. Specifically, the derivative of the log-likelihood of $\bs{x}\in\mathcal{X}$ w.r.t. $\bs{\theta}$ can be expressed as
\begin{align}\label{eq:ml}
  \frac{\partial\log p_{\bs{\theta}}(\bs{x})}{\partial\bs{\theta}}\!=\!   \mathbb{E}_{p_{\bs{\theta}}(\bs{x})}\!\!\left[ \frac{\partial E_{\bs{\theta}}(\bs{x})}{\partial\bs{\theta}} \right]
  \!-\!\mathbb{E}_{p_d(\bs{x})}\!\!\left[ \frac{\partial E_{\bs{\theta}}(\bs{x})}{\partial\bs{\theta}} \right],
\end{align}
where $p_d(\bs{x})$ is the real data distribution (i.e., training dataset), and $p_{\bs{\theta}}(\bs{x})$ is the estimated probability density function, sampling from which is challenging due to the intractable $Z(\bs{\theta})$. 

Prior works have developed a number of sampling-based approaches to sample from $p_{\bs{\theta}}(\bs{x})$ efficiently, such as MCMC and Gibbs sampling~\cite{hinton2002cd}. By utilizing the gradient information, Stochastic Gradient Langevin Dynamics (SGLD)~\cite{welling2011bayesian} has been employed recently to speed up the sampling from $p_{\bs{\theta}}(\bs{x})$~\cite{nijkamp2019learning,du2019implicit,jem}. Specifically, to sample from $p_{\bs{\theta}}(\bs{x})$, the SGLD follows
\begin{align}\label{eq:sgld}
    &\bs{x}^0\sim p_0(\bs{x}),  \nonumber\\
    &\bs{x}^{t+1} = \bs{x}^t-\frac{\alpha}{2} \frac{\partial
    E_{\bs{\theta}}(\bs{x}^t)}{\partial \bs{x}^t} + \alpha\bs{\epsilon}^t, \;\;
    \bs{\epsilon}^t \sim \mathcal{N} (\bs{0},\bs{1}),
\end{align}
where $p_0(\bs{x})$ is typically a uniform distribution over $[-1,1]$, whose samples are refined via a noisy gradient decent with step-size $\alpha$ over a sampling chain. 

\paragraph{Joint Energy-based Model} (JEM)~\cite{jem} reinterprets the standard softmax classifier as an EBM and trains a single network for hybrid discriminative-generative modeling. Specifically, Grathwohl et al.~\cite{jem} were the first to recognize the logits $f_{\bs{\theta}}(\bs{x})[y]$ from a standard softmax classifier can be considered as an energy function over $(\bs{x}, y)$, and thus the joint density can be defined as $p_{\bs{\theta}}(\bs{x}, y)=e^{f_{\bs{\theta}}(\bs{x})[y]} / Z(\bs{\theta})$, where $Z(\bs{\theta})$ is an unknown normalizing constant (regardless of $\bs{x}$ or $y$). Then the density of $\bs{x}$ can be derived by marginalizing over $y$: $p_{\bs{\theta}}(\bs{x})=\sum_{y} p_{\bs{\theta}}(\bs{x}, y) = \sum_{y} e^{f_{\bs{\theta}}(\bs{x})\left[y\right]} / Z(\bs{\theta})$. Subsequently, the corresponding energy function of $\bs{x}$ can be identified as 
\begin{equation}\label{eq:jem_ex}
    E_{\bs{\theta}} (\bs{x})\!=\!-\log\! \sum_{y}\!e^{f_{\bs{\theta}}(\bs{x})\left[y\right]}\!=\!-\text{LSE}( f_{\bs{\theta}}(\bs{x})),
\end{equation}
where $\text{LSE}(\cdot)$ denotes the Log-Sum-Exp function.

To optimize the model parameter $\bs{\theta}$, JEM maximizes the logarithm of joint density function $p_{\bs{\theta}}(\bs{x},y)$:
\begin{equation}\label{eq:jem_loss}
  \log p_{\bs{\theta}}(\bs{x}, y) = \log p_{\bs{\theta}}(y|\bs{x}) + \log p_{\bs{\theta}}(\bs{x}),
\end{equation}
where the first term denotes the cross-entropy objective for classification, and the second term can be optimized by the maximum likelihood learning of EBM as shown in Eq.~\ref{eq:ml}.

However, JEM suffers from high training instability even with a large number of SGLD sampling steps $K$ (e.g., $K=20$). After divergence, JEM requires to restart the SGLD sampling with a doubled $K$. Recently, JEM++~\cite{jempp} proposes a number of new training techniques to improve JEM’s accuracy, training stability and speed altogether, including the proximal gradient clipping, YOPO-based SGLD sampling acceleration, and informative initialization. Furthermore, JEM++ enables batch norm~\cite{batchnorm15} in the backbone models, while IGEBM and JEM have to exclude batch norm due to the high training instability incurred by it. 

\paragraph{Flat Minima and Generalization}
 
A great number of previous works have investigated the relationship between the flatness of local minima and the generalization of learned models~\cite{losslandscape,keskar2016large,wei2020implicit,vitsam,sam2021,asam2021}. Now it is widely accepted and empirically verified that flat minima tend to give better generalization performance. Based on these observations, several recent regularization techniques are proposed to search for the flat minima of loss landscapes~\cite{wei2020implicit,vitsam,sam2021,asam2021}. Among them, the sharpness-aware minimization (SAM)~\cite{sam2021} is a recently introduced optimizer that demonstrates promising performance across all kinds of models and tasks, such as ResNet~\cite{resnet16}, Vision Transformer (ViT)~\cite{vitsam} and Language Modeling~\cite{samlm}. Furthermore, score matching-based methods~\cite{hyvarinen2005estimation,swersky2011autoencoders,song2019generative,song2020score} also explore the behaviour of flat minima in generative models and learn unnormalized statistical models by matching the gradient of the log probability density of the model distribution to that of the data distribution. To the best of our knowledge, we are the first to explore the sharpness-aware optimization to improve both the discriminative and generative performance of EBMs.

\section{SADA-JEM}
\subsection{Sharpness-Aware Minimization}

To train a generalizable model, SAM~\cite{sam2021} proposes to search for model parameters $\bs{\theta}$ whose entire neighborhoods have uniformly low loss values by optimizing a minimax objective:
\begin{align}\label{eq:sam_obj}
    \min_{\bs{\theta}}\ \max_{\| \bs{\epsilon} \|_2\leq \rho} L_{train}(\bs{\theta}+ \bs{\epsilon}) + \lambda \| \bs{\theta} \|^2_2,
\end{align}
where $\rho$ is the radius of the $L_2$-ball centered at model parameters $\bs{\theta}$, and $\lambda$ is a hyperparameter for $L_2$ regularization on $\bs{\theta}$. To solve the inner maximization problem, SAM employs the Taylor expansion to develop an efficient first-order approximation to the optimal $\bs{\epsilon}^*$ as:
\begin{align}\label{eq:e_theta}
    \hat{\bs{\epsilon}}(\bs{\theta}) & = \argmax_{\|\epsilon\|_2\leq\rho}L_{train}(\bs{\theta}) + \epsilon^T\nabla_{\bs{\theta}} L_{train}(\bs{\theta}) \nonumber\\
    & = \rho\nabla_{\bs{\theta}} L_{train}(\bs{\theta})/\|\nabla_{\bs{\theta}} L_{train}(\bs{\theta})\|_2,
\end{align}
which is a scaled $L_2$ normalized gradient at the current model parameters $\bs{\theta}$. Once $\hat{\bs{\epsilon}}$ is determined, SAM updates $\bs{\theta}$ based on the gradient $\nabla_{\bs{\theta}} L_{train}(\bs{\theta})|_{\bs{\theta}+\hat{\epsilon}(\bs{\theta})}+2\lambda\bs{\theta}$ at an updated parameter location $\bs{\theta}+\hat{\epsilon}$. More recently, Kwon et al.~\cite{asam2021} propose an Adaptive SAM (ASAM) with the objective:\vspace{-5pt}
\begin{align}\label{eq:asam_obj}
    \min_{\bs{\theta}}\ \max_{\|T^{-1}_{\bs{\theta}} \bs{\epsilon} \|_2\leq \rho} L_{train}(\bs{\theta}+ \bs{\epsilon}) + \lambda \| \bs{\theta} \|^2_2,
\end{align}
where $T_{\bs{\theta}}$ is an element-wise operator $T_{\bs{\theta}}\!=\!\text{diag}(|\theta_{1}|,|\theta_{2}|, \dots,|\theta_{k}|)$ with $\bs{\theta}\!=\![\theta_{1}, \theta_{2},\dots, \theta_{k}]$. Similar to SAM, the Taylor expansion is leveraged in ASAM to derive a first-order approximation to the optimal $\bs{\epsilon}^*$ with $\hat{\bs{\epsilon}}(\bs{\theta}) = \rho\,T_{\bs{\theta}}\,\text{sign} (\nabla L_{train}(\bs{\theta}))$. 

As we observed from Figure~\ref{figure:energy_landscape}(a) and (b), models trained by JEM converge to very sharp local optima, which potentially undermines the generalization of JEM. We therefore incorporate the framework of SAM to the original training pipeline of JEM~\cite{jem} in order to improve the generalization of trained models. Specifically, instead of the traditional maximum likelihood training, we optimize the joint density function of JEM in a minimax objective:
\begin{align}\label{eq:sajem_obj}
    \max_{\bs{\theta}}\ \min_{\| \bs{\epsilon} \|_2\leq \rho} \log p_{(\bs{\theta}+ \bs{\epsilon})}(\bs{x},y) + \lambda \| \bs{\theta} \|^2_2.
\end{align}
For the outer maximization that involves $\log p_{\bs{\theta}}(\bs{x})$, SGLD is again used to sample from $p_{\bs{\theta}}(\bs{x})$ as in the original JEM.


\subsection{Image Generation without Data Augmentation}
Data augmentation is a critical technique in supervised deep learning and self-supervised contrastive learning~\cite{imagenet12,simclr}. Not surprisingly, JEM also utilizes data augmentation in its training pipeline, such as horizontal flipping, random cropping, and padding. Specifically, let $T$ denote a data augmentation operator. The actual objective function of JEM is
\begin{equation}
  \log p_{\bs{\theta}}(\bs{x}, y) = \log p_{\bs{\theta}}(y|T(\bs{x})) + \log p_{\bs{\theta}}(T(\bs{x})),
\end{equation}
which shows that JEM maximizes the likelihood function $p_{\bs{\theta}}(T(\bs{x}))$ rather than $p_{\bs{\theta}}(\bs{x})$. 
From our empirical studies, horizontal flipping has little impact on the image generation quality, while cropping and padding play a bigger role because the generated images contain cropping and padding effects, which hurt the quality of generated images. This is consistent with GANs~\cite{GAN},  which observed that any augmentation that is applied to the training dataset will get inherited in the generated images. Based on this observation, we exclude the data augmentation from $p_{\bs{\theta}}(T(\bs{x}))$ and only retain the data augmentation for classification given its pervasive success in image classification. To this end, our final objective function of SADA-JEM becomes:
\begin{equation}\label{eq:obj_jem}
  \log p_{\bs{\theta}}(\bs{x}, y) = \log p_{\bs{\theta}}(y|T(\bs{x})) + \log p_{\bs{\theta}}(\bs{x}),
\end{equation}
where the first term is calculated using a mini-batch with data augmentation, and the second term is calculated using a mini-batch without data augmentation, which can be implemented efficiently by using two data loaders. StyleGAN2-ADA~\cite{styleganADA} proposes a type of ``non-leaking" data augmentation to prevent the discriminator from overfitting, and thus improves the image quality. However, from our empirical studies, we find that this technique hurts the performance of both image quality and classification accuracy.

\begin{algorithm}[t]
\caption{SADA-JEM Training: Given network $f_\theta$, SGLD step-size $\alpha$, SGLD noise $\sigma$, SGLD steps $K$, replay buffer $B$, reinitialization frequency $\gamma$, SAM noise bound $\rho$, and learning rate $lr$}
\label{algo:1}
\begin{algorithmic}[1]
\WHILE{not converged}
\STATE Sample $\bs{x}^+$ and $y$ from training dataset
\STATE Sample $\widehat{\bs{x}}_0 \sim B$ with probability $1-\gamma$,  else $\widehat{\bs{x}}_0 \sim p_0(\bs{x})$  
\FOR{$t \in [1, 2, \ldots, K]$}   
  \STATE $\widehat{\bs{x}}_t = \widehat{\bs{x}}_{t-1} - \alpha \cdot \frac{\partial E(\widehat{\bs{x}}_{t-1})}{\partial \widehat{\bs{x}}_{t-1}} + \sigma \cdot \mathcal{N}(0, I)$
\ENDFOR
\STATE $\bs{x}^- = \text{StopGrad}(\widehat{\bs{x}}_K)$
\STATE $L_{\text{gen}}(\theta) = E(\bs{x}^+) - E(\bs{x}^-)$ \label{row:cd}
\STATE $L(\bs{\theta}) = L_\text{clf}(\bs{\theta}) +  L_{\text{gen}}(\bs{\theta})$ with  $L_{\text{clf}}(\bs{\theta}) = \text{xent}(f_{\bs{\theta}}(\bs{x}), y)$ \label{row:full}
\STATE \# Apply SAM optimizer as following:
\STATE Compute gradient $\nabla_{\bs{\theta}}{L(\bs{\theta})}$ of the training loss
\STATE Compute $\hat{\bs{\epsilon}}(\bs{\theta})$ with $\rho$ as in Eq.~\ref{eq:e_theta}
\STATE Compute gradient $\bs{g} = \nabla_{\bs{\theta}}{L(\bs{\theta})}|_{\bs{\theta} + \hat{\bs{\epsilon}(\bs{\theta})}}$
\STATE Update model parameters: $\bs{\theta} = \bs{\theta} - lr \cdot  \bs{g}$
\STATE Push $\bs{x}^-$ to $B$
\ENDWHILE
\end{algorithmic}
\end{algorithm}

Algorithm~\ref{algo:1} provides the pseudo-code of SADA-JEM training, which follows a similar design of JEM~\cite{jem} and JEM++~\cite{jempp} with a replay buffer. For brevity, only one real sample and one generated sample are used to optimize model parameters $\bs{\theta}$. But it is straightforward to generalize the pseudo-code below to a mini-batch setting, which we use in the experiments. It is worth mentioning that we adopt the Informative Initialization in JEM++ to initialize the Markov chain from $p_0(\bs{x})$, which enables the batch norm and plays a crucial role in the tradeoff between the number of SGLD sampling steps $K$ and overall performance, including the classification accuracy and training stability.


\section{Experiments}
We train SADA-JEM with the Wide-ResNet 28-10~\cite{wideresnet16} backbone on CIFAR10 and CIFAR100, and evaluate its performance on a set of discriminative and generative tasks, including image classification, generation, calibration, out-of-distribution (OOD) detection, and adversarial robustness. Our code is built on top of JEM++~\cite{jempp}\,\footnote{\url{https://github.com/sndnyang/jempp}} (given its improved performance over JEM) and SAM\,\footnote{\url{https://github.com/davda54/sam}}. For a fair comparison, our experiments largely follow the settings of JEM and JEM++, with details provided in the supplementary material. All our experiments are conducted using PyTorch on a single Nvidia RTX GPU. 

\subsection{Hybrid Modeling}

\begin{figure}[ht]
    \centering
    \subfloat[CIFAR10]{
        \includegraphics[width=0.3\columnwidth]{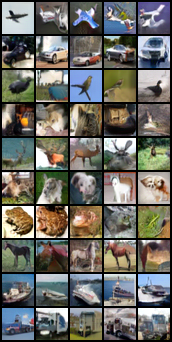}
        \label{figure:CIFAR10_con_sample}
    }
    \subfloat[CIFAR100]{
        \includegraphics[width=0.48\columnwidth]{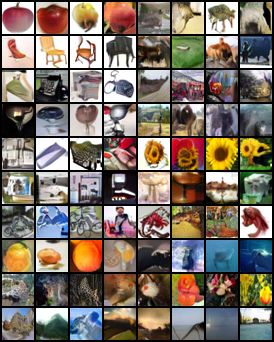}
        \label{figure:CIFAR100_con_sample}
    }
    \vspace{-5pt}
    \caption{Generated samples from SADA-JEM.}
    \label{figure:sajem_samples}
    \vspace{-10pt}
\end{figure}

We first compare the performance of SADA-JEM with state-of-the-art hybrid models, stand-alone discriminative models, and generative models on CIFAR10 and CIFAR100, with the results reported in  Table~\ref{table:hybrid_results} and~\ref{table:hybrid_cifar100}. Inception Score (IS)~\cite{imprgan16} and Fr\'{e}chet Inception Distance (FID)~\cite{heusel2017gans} are employed to measure the quality of generated images. It can be observed from Table~\ref{table:hybrid_results} that SADA-JEM ($K\!=\!5$) outperforms JEM ($K\!=\!20$) and JEM++ ($M\!=\!20$) in classification accuracy (95.5\%) and the FID score (9.41) on CIFAR10, where the FID score of SADA-JEM is a dramatic improvement over that of JEM/JEM++'s (37.1). Similarly, Table~\ref{table:hybrid_cifar100} shows that the improvement of SADA-JEM over JEM/JEM++ on CIFAR100 is also significant: the FID score is improved from 33.7 to 14.4. Moreover, we find that SADA-JEM is superior in training stability too. For instance, SADA-JEM ($K\!=\!5$) outperforms JEM++ ($M\!=\!20$) in classification accuracy, while exhibiting a much higher training stability than JEM/JEM++\,\footnote{JEM ($K\!=\!20$) and JEM++ ($M\!=\!5$) can easily diverge at early epochs.}. Example images generated by SADA-JEM for CIFAR10 and CIFAR100 are provided in Figure~\ref{figure:sajem_samples}. 

\begin{table}[ht]
\caption{Results on CIFAR10}\vspace{-15pt}
\label{table:hybrid_results}
\begin{center}
\begin{threeparttable}
\begin{tabular}{lccc}
\toprule
 Model                                      & Acc \% $\uparrow$ & IS $\uparrow$ & FID $\downarrow$ \\
 \midrule
SADA-JEM (K=5)                                  & 95.5 & 8.77 & 9.41    \\
SADA-JEM (K=10)                                 & 96.0 & 8.63 & 11.4 \\
SADA-JEM (K=20)                                 & 96.1 & 8.40 & 13.1 \\
\midrule
    \multicolumn{4}{c}{Single Hybrid Model} \\
IGEBM (K=60)~\cite{du2019implicit}           & 49.1 & 8.30 & 37.9 \\
JEM (K=20)*~\cite{jem}                        & 92.9 & 8.76 & 38.4 \\
JEM++ (M=5)*~\cite{jempp}                       & 91.1 & 7.81 & 37.9  \\
JEM++ (M=10)~\cite{jempp}                                & 93.5 & 8.29 & 37.1  \\
JEM++ (M=20)~\cite{jempp}                                & 94.1 & 8.11 & 38.0  \\
JEAT~\cite{jeat}                            & 85.2 & 8.80 & 38.2  \\
\midrule
\multicolumn{4}{c}{Other EBMs} \\ 
CF-EBM (K=50)~\cite{cfebm}                   & - &   -    & 16.7 \\
ImCD (K=40)~\cite{improvedCD}                & - &  7.85  & 25.1 \\
DiffuRecov (K=30)~\cite{diffusionRecovery}   & - &  8.31  &  9.58 \\
VAEBM (K=6)~\cite{vaebm}                     & - &  8.43  & 12.2 \\
VERA~\cite{nomcmc}                          & 93.2 & 8.11 & 30.5 \\
\midrule 
\multicolumn{4}{c}{Other Models} \\
 Softmax                                    & 96.2 & - & - \\
 Softmax + SAM                              & \textbf{97.2} & - & - \\
 SNGAN~\cite{miyato2018spectral}            & - & 8.59  & 21.7  \\
 StyleGAN2-ADA~\cite{styleganADA}                              & - & \textbf{9.74}  & \textbf{2.92}  \\
\bottomrule
\end{tabular}
\begin{tablenotes}
  \scriptsize\item * The training is unstable and regularly diverged.
\end{tablenotes}
\end{threeparttable}
\end{center}
\vspace{-5pt}
\end{table}

\begin{table}[ht!]
\caption{Results on CIFAR100}
\label{table:hybrid_cifar100}
\vspace{-15pt}
\begin{center}
\begin{threeparttable}
\begin{tabular}{l|ccc}
\toprule
Model                              &  Acc \% $\uparrow$ & IS $\uparrow$ & FID $\downarrow$  \\
\midrule
SADA-JEM (K=5)                         & 75.0     & \textbf{11.63}  & 14.4 \\
SADA-JEM (K=10)                        & 76.4     & 10.95  & 15.1 \\
SADA-JEM (K=20)                        & 77.3     & 10.78  & 19.9 \\
\midrule
JEM (K=20)*~\cite{jem}                         & 72.2 & 10.22 & 38.1  \\
JEM++ (M=5)*~\cite{jempp}                        & 72.1 &  8.05 & 38.9  \\
JEM++ (M=10)*~\cite{jempp}                       & 74.2 &  9.97 & 34.5  \\
JEM++ (M=20)*~\cite{jempp}                       & 75.9 & 10.07 & 33.7  \\
VERA ($\alpha$=100)*~\cite{nomcmc}                & 72.2 & 8.25  & 29.5  \\
VERA ($\alpha$=1)*~\cite{nomcmc}                  & 48.7 & 7.84  & 25.1  \\
\midrule
Softmax                            & 81.3 &  -   &  -    \\
Softmax + SAM                      & \textbf{83.4} &  -   &  -    \\
SNGAN~\cite{miyato2018spectral}                              & -    & 9.30 & 15.6  \\
BigGAN~\cite{biggan}                             & -    & 11.0 & \textbf{11.7}  \\
\bottomrule
\end{tabular}
\begin{tablenotes}
  \scriptsize\item * No official IS and FID scores are reported. We run the official code with the default settings and report the results.
\end{tablenotes}
\end{threeparttable}
\end{center}
   \vspace{-10pt}
\end{table}

\begin{figure}
  \centering
  \begin{subfigure}{0.48\linewidth}
    \includegraphics[width=1\columnwidth]{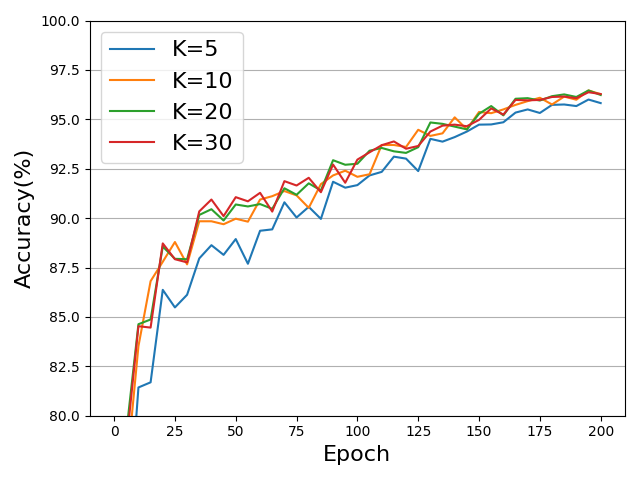}
    \caption{Accuracy}
    \label{figure:K_acc}
  \end{subfigure}
  \hfill
  \begin{subfigure}{0.48\linewidth}
    \includegraphics[width=1\columnwidth]{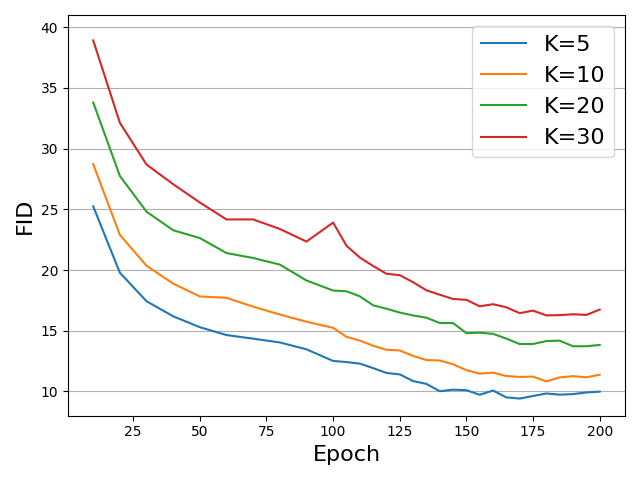}
    \caption{FID (the lower the better)}
    \label{figure:K_FID}
  \end{subfigure}\vspace{-5pt}
    \caption{The learning curves of SADA-JEM on CIFAR10 with different SGLD sampling steps $K$.}
    \label{figure:K_curves}
    \vspace{-10pt}
\end{figure}

One interesting phenomenon we observed from our experiments is that the image quality often drops as number of SGLD sampling steps $K$ increases, as shown in Figure~\ref{figure:K_curves}(b). A similar observation has been reported in IGEBM~\cite{du2019implicit}, where the authors found that a large $K$ can facilitate the convergence of SGLD to high likelihood modes of an energy landscape, but often leads to saturated images and thus degraded image quality. Unlike IGEBM, SADA-JEM is a hybrid model that trains one single network for image classification and generation. As we can see from Figure~\ref{figure:K_curves}, as $K$ increases the classification accuracy of SADA-JEM increases (insignificantly), while the image quality drops. Therefore, it seems there is a performance trade-off between classification accuracy and image generation quality, and SADA-JEM's performances on both tasks are not always positively correlated after certain points (e.g., $K$). This is an interesting observation that we believe is worthy of further investigation.


\begin{figure*}[ht!]
  \centering
  \begin{subfigure}{0.24\linewidth}
    \includegraphics[width=1\columnwidth]{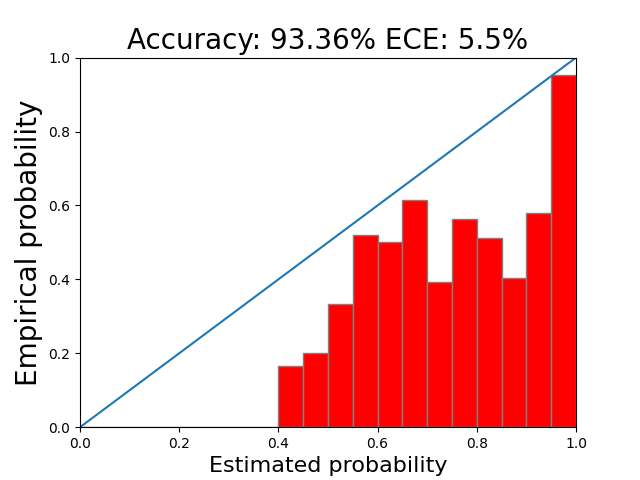}
    \caption{Softmax (w/o BN)}
    \label{figure:base_cali}
  \end{subfigure}
  \hfill
  \begin{subfigure}{0.24\linewidth}
    \includegraphics[width=1\columnwidth]{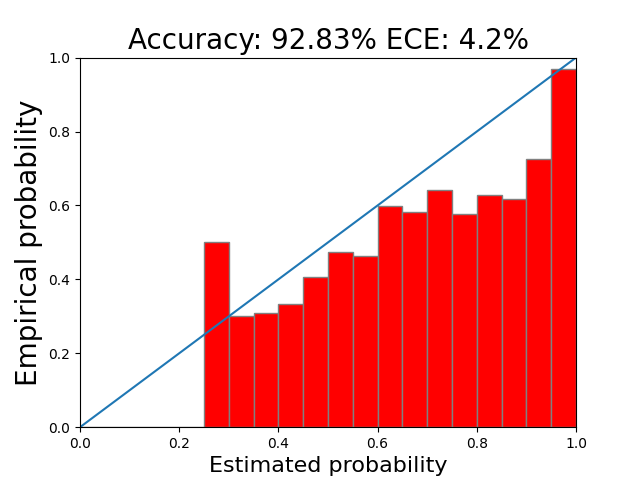}
    \caption{JEM (K=20)}
    \label{figure:jem_cali}
  \end{subfigure}
  \hfill
  \begin{subfigure}{0.24\linewidth}
    \includegraphics[width=1\columnwidth]{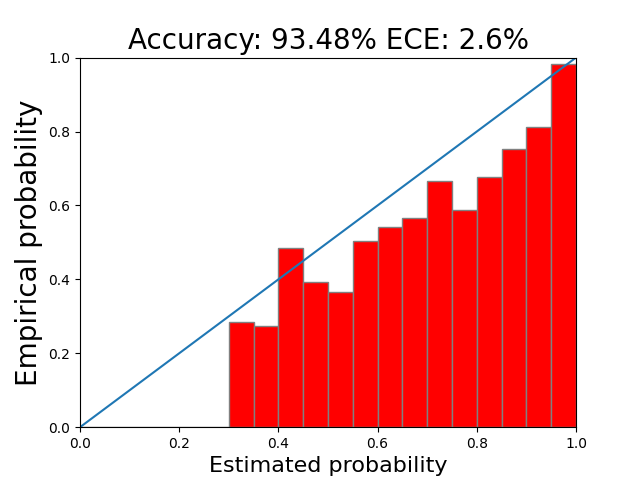}
    \caption{JEM++ (M=10)}
    \label{figure:jempp_M10_cali}
  \end{subfigure}
  \hfill
  \begin{subfigure}{0.24\linewidth}
    \includegraphics[width=1\columnwidth]{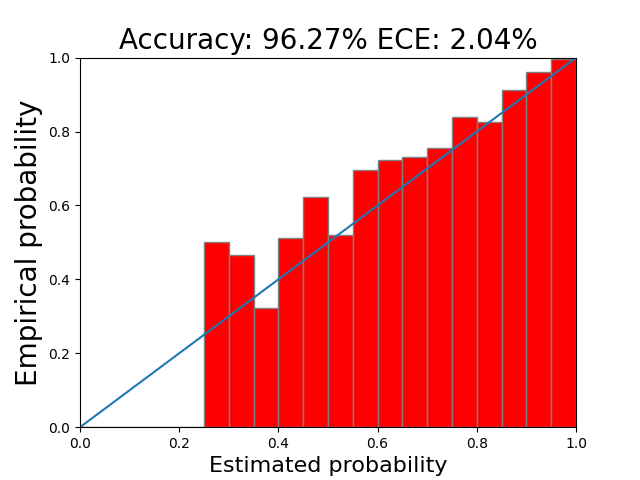}
    \caption{SADA-JEM (K=10)}
    \label{figure:sajem_K10_cali}
  \end{subfigure}\vspace{-5pt}
    \caption{Calibration results on CIFAR10. The smaller ECE is, the better.}
    \label{figure:CIFAR10_cali}
\end{figure*}

\subsection{Calibration}

While modern classifiers are growing more accurate, recent works show that their predictions could be over-confident due to increased model capacity~\cite{guo2017calibration}. Typically, the confidence of a model's prediction can be defined as $\max_y p(y|\bs{x})$ and is used to decide whether to output a prediction or not. However, incorrect but confident predictions can be catastrophic for safety-critical applications, which necessitates calibration of uncertainty especially for models of large capacity. As such, a well-calibrated but less accurate model can be considerably more useful than a more accurate but less-calibrated model.

In this experiment, all models are trained on the CIFAR10 dataset for a fair comparison. We compare the Expected Calibration Error (ECE) score~\cite{guo2017calibration} of SADA-JEM to those of the standard softmax classifier and JEM. We utilize the \textit{reliability diagram} to visualize the discrepancy between the true probability and the confidence, with the results shown in Figure~\ref{figure:CIFAR10_cali}. We find that the model trained by SADA-JEM ($K=10$) achieves a much smaller ECE (2.04\% vs. 4.2\% of JEM and 5.5\% of softmax classifier), demonstrating SADA-JEM's predictions are better calibrated than the competing methods. Similar to image quality, we notice that a larger $K$ also undermines the calibration quality slightly. Due to page limit, more results are relegated to the supplementary material.

\begin{table*}[ht!]
\caption{OOD detection results. Models are trained on CIFAR10. Values are AUROC.}
\vspace{-15pt}
\label{table:CIFAR10_ood}
\begin{center}
\begin{threeparttable}
\begin{tabular}{c|c|cccc}
\toprule
$s_{\bs{\theta}}(\bs{x})$  & Model   & SVHN & CIFAR10 Interp & CIFAR100 & CelebA \\
\midrule
\multirow{9}{*}{$\log p_{\bs{\theta}}(\bs{x})$} & WideResNet~\cite{liu2020energy_ood}          & .91 & - & .87 & .78 \\
            & IGEBM~\cite{du2019implicit}    & .63 & .70 & .50 & .70 \\
            & JEM (K=20)~\cite{jem}          & .67 & .65 & .67 & .75 \\
            & JEM++ (M=20)~\cite{jempp}      & .85 & .57 & .68 & .80 \\
            & VERA~\cite{nomcmc}             & .83 & \bf{.86} & .73 & .33 \\
            & ImCD~\cite{improvedCD}       & .91 & .65 & .83 & - \\
            & SADA-JEM (K=5)           & .91 & .79 & .90 & .82 \\
            & SADA-JEM (K=10)          & .95 & .81 & .90 & .88 \\
            & SADA-JEM (K=20)          & \bf{.98} & .83 & \bf{.92} & \bf{.95} \\
\midrule
\multirow{7}{*}{$\max_y p_{\bs{\theta}}(y|\bs{x})$} & WideResNet             & .93 & .77 & .85 & .62 \\
    & IGEBM~\cite{du2019implicit}        & .43 & .69 & .54 & .69 \\
    & JEM (K=20)~\cite{jem}              & .89 & .75 & .87 & .79 \\
    & JEM++ (M=20)~\cite{jempp}          & .94 & .77 & .88 & \bf{.90} \\
    & SADA-JEM (K=5)            & .92 & .77 & .88 & .81 \\
    & SADA-JEM (K=10)           & .93 & .78 & .89 & .78 \\
    & SADA-JEM (K=20)           & \bf{.96} & \bf{.80} & \bf{.91} & .84 \\
\bottomrule
\end{tabular}
\end{threeparttable}
\end{center}
\vspace{-10pt}
\end{table*}

\subsection{Out-Of-Distribution Detection}

Formally, the OOD detection is a binary classification problem, which outputs a score $s_{\bs{\theta}}(\bs{x}) \in \mathbb{R}$ for a given query $\bs{x}$. The model should be able to assign lower scores to OOD examples than to in-distribution examples such that it can be used to distinguish OOD examples from in-distribution ones. Following the settings of JEM~\cite{jem}, we use the Area Under the Receiver-Operating Curve (AUROC)~\cite{HenGim16} to evaluate the performance of OOD detection. In our experiments, two score functions are considered: the input density $p_{\bs{\theta}}(\bs{x})$~\cite{nalisnick2018deep}, and the predictive distribution $p_{\bs{\theta}}(y|\bs{x})$~\cite{HenGim16}.

\vspace{-5pt}
\paragraph{Input Density}
We can use the input density $p_{\bs{\theta}}(\bs{x})$ as $s_{\bs{\theta}}(\bs{x})$. Intuitively, examples with low $p(\bs{x})$ are considered to be OOD samples. Quantitative results can be found in Table~\ref{table:CIFAR10_ood} (top row), where CIFAR10 is the in-distribution data, and SVHN, an interpolated CIFAR10, CIFAR100 and CelebA are the out-of-distribution data, respectively. Moreover, the corresponding visualization are shown in Table~\ref{table:logpx_hist}. As we can see, SADA-JEM performs better in distinguishing the in-distribution samples from OOD ones, outperforming JEM, JEM++ and most of the other models by significant margins.

\vspace{-5pt}
\paragraph{Predictive Distribution}
Another useful OOD score function is the maximum probability from a classifier's predictive distribution: $s_{\bs{\theta}}(\bs{x}) = \max_y p_{\bs{\theta}}(y|\bs{x})$. Hence, OOD performance using this score is highly correlated with a model's classification accuracy. Table~\ref{table:CIFAR10_ood} (bottom row) reports the results of this method. Again, SADA-JEM outperforms JEM and all the other models in majority of cases.


Table~\ref{table:CIFAR10_ood} (top row) also shows that JEM and JEM++ have even worse performance than a standard classifier in OOD detection. This is likely because both JEM and JEM++ maximize $p_{\bs{\theta}}(T(\bs{x}))$ with data augmentation $T$, which undesirably enlarges the span of estimated $p_{\bs{\theta}}(\bs{x})$ and makes it less distinguishable to the OOD samples. In contrast, VERA, ImCD, and SADA-JEM exclude the data augmentation from their training pipelines, and consistently, they all demonstrate improved OOD detection performance over JEM and JEM++.

\begin{table*}[ht!]
  \centering
  \begin{tabular}{ | c | m{4.1cm} | m{4.1cm} | m{4.1cm} | }
    \hline
    JEM
    &
    \begin{minipage}{.23\textwidth}
      \includegraphics[width=\linewidth, height=32mm]{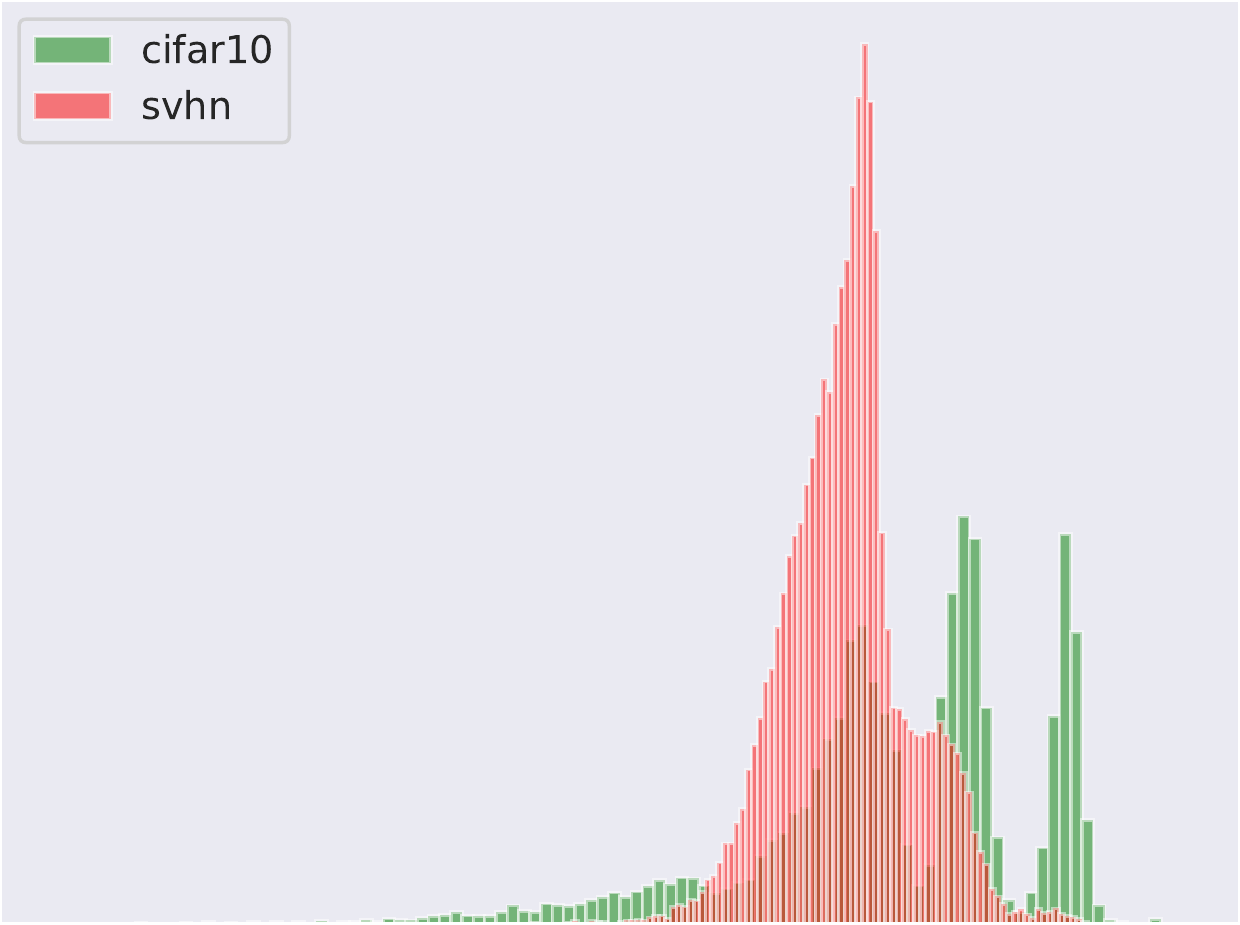}
    \end{minipage}
    &
    \begin{minipage}{.23\textwidth}
      \includegraphics[width=\linewidth, height=32mm]{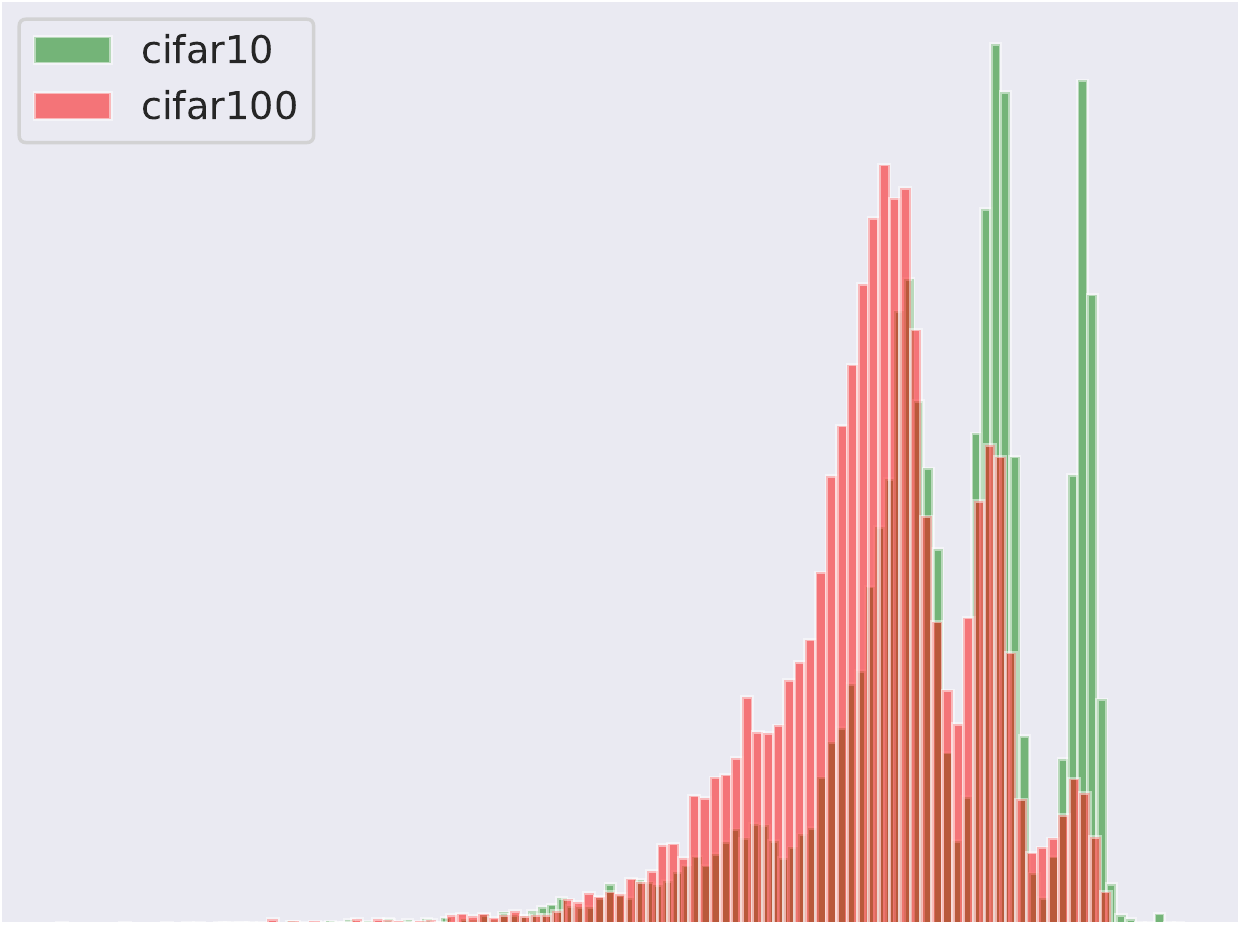}
    \end{minipage}
    &
    \begin{minipage}{.23\textwidth}
      \includegraphics[width=\linewidth, height=32mm]{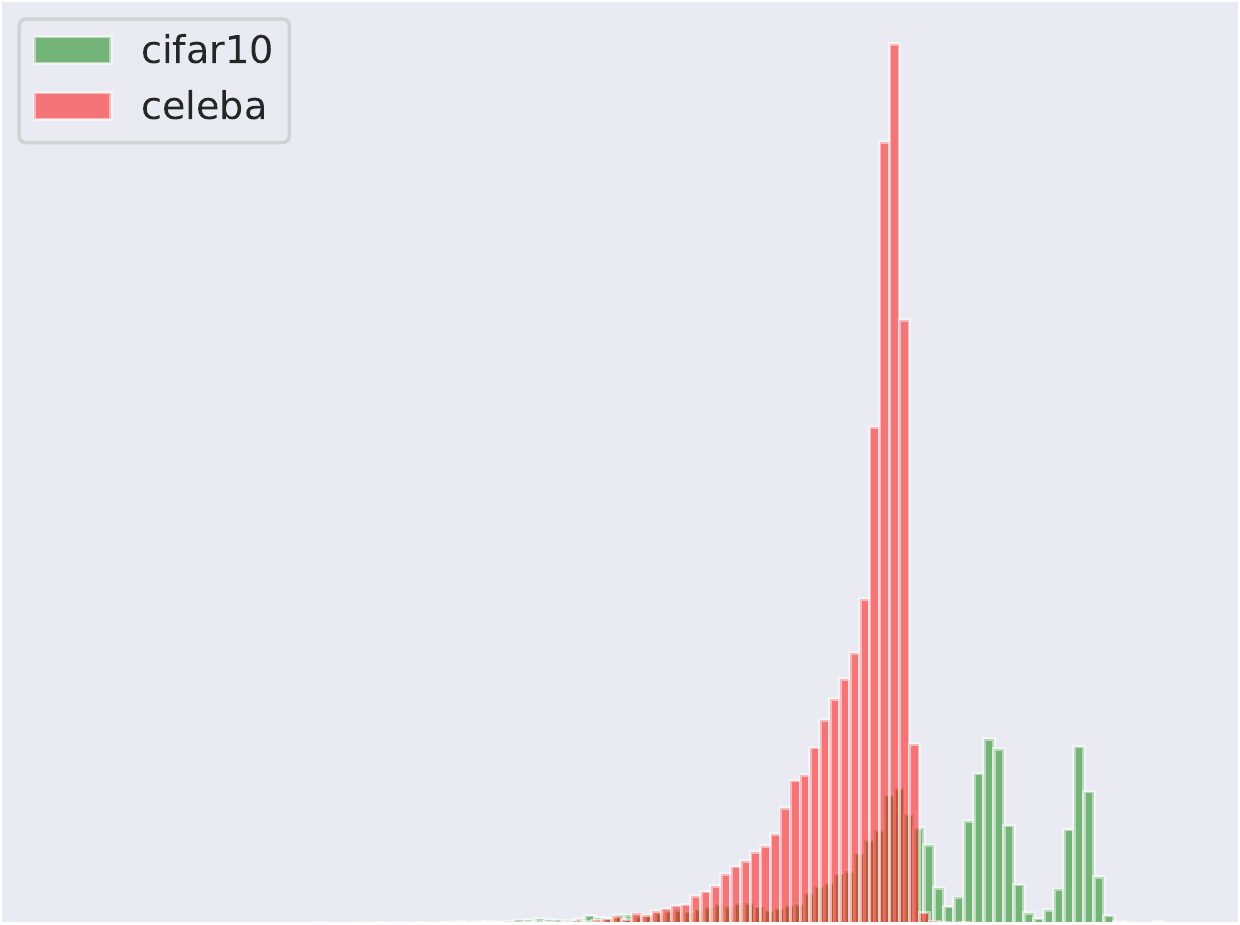}
    \end{minipage}
      \\ \hline
    JEM++ (M=10)
    &
    \begin{minipage}{.23\textwidth}
      \includegraphics[width=\linewidth, height=32mm]{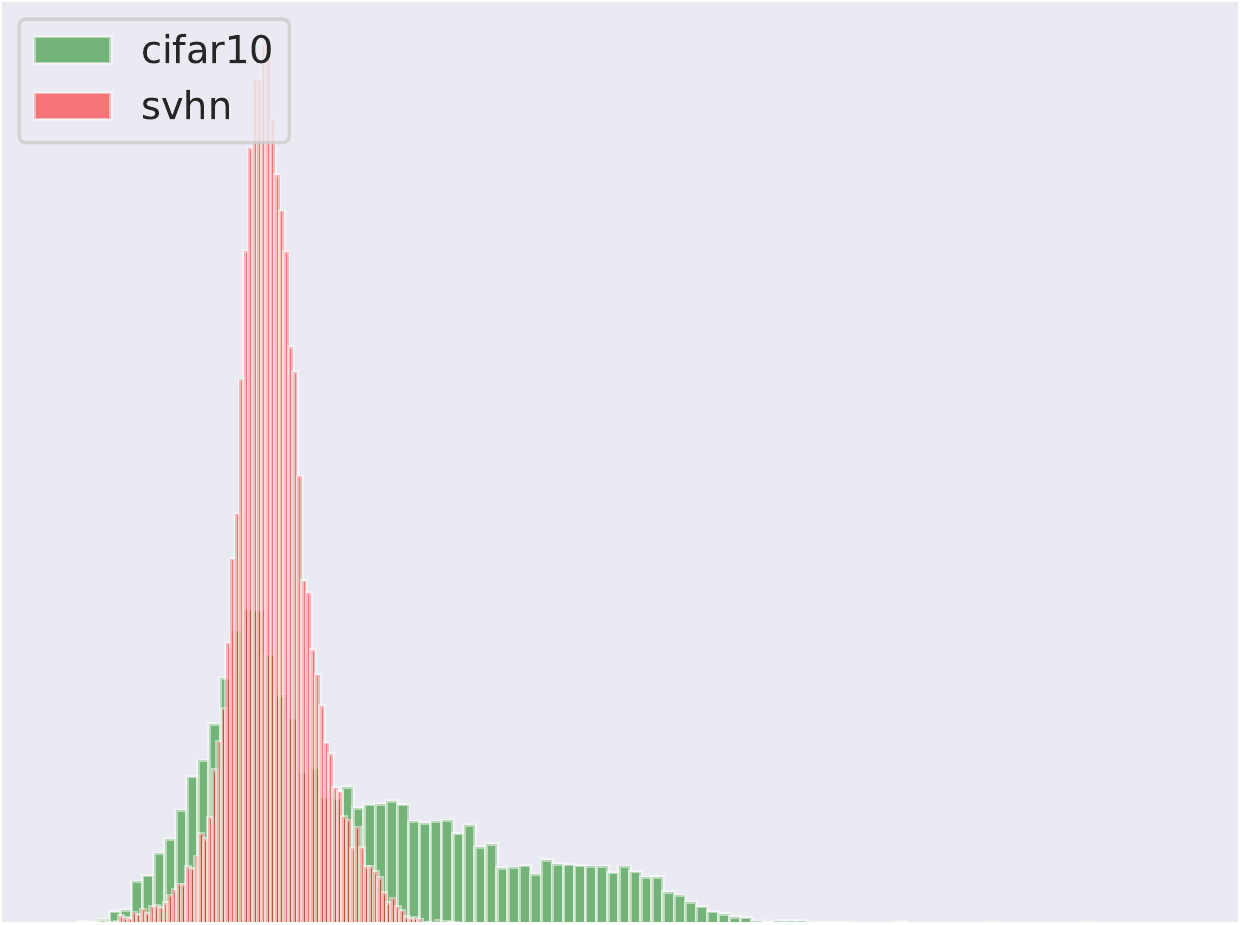}
    \end{minipage}
    &
    \begin{minipage}{.23\textwidth}
      \includegraphics[width=\linewidth, height=32mm]{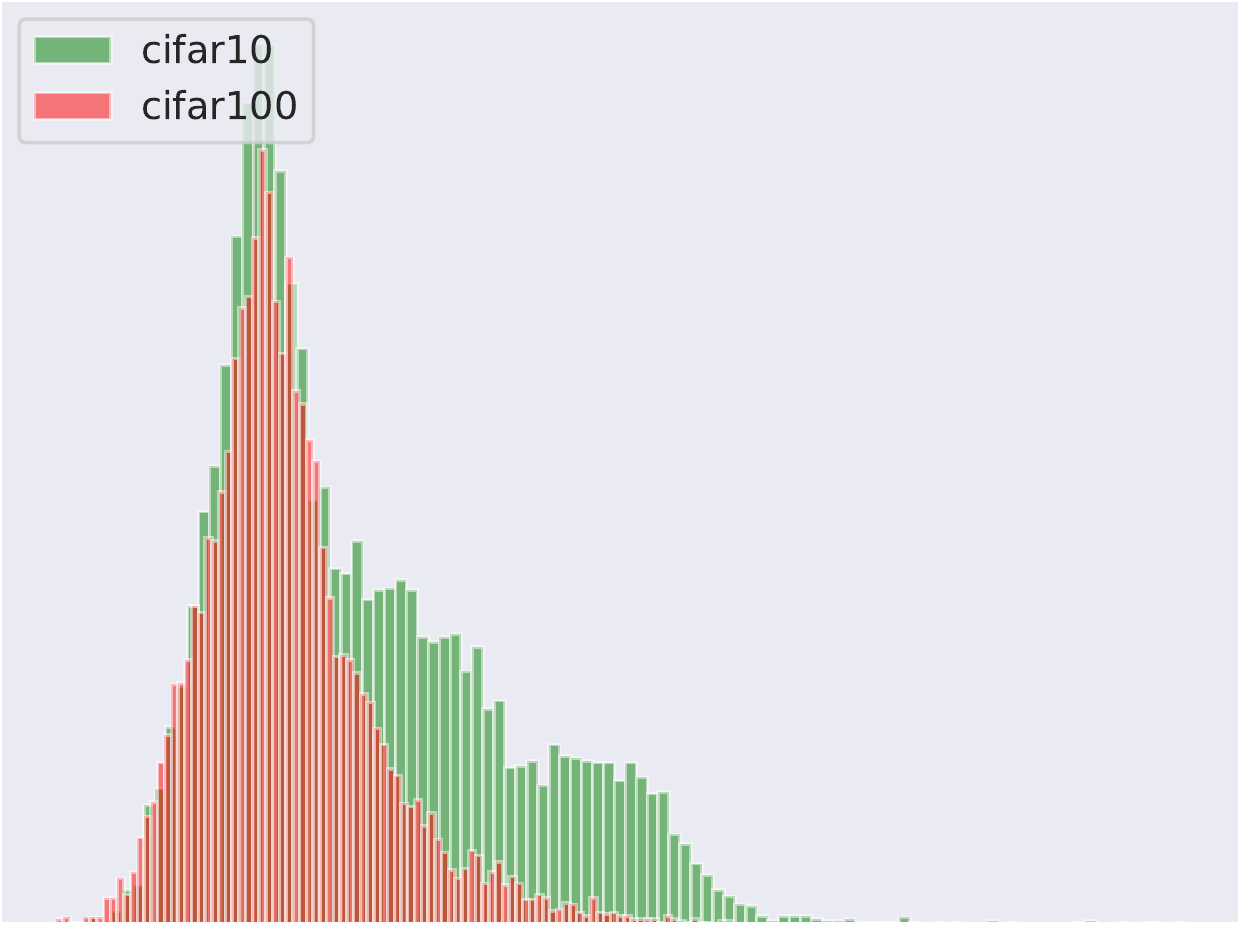}
    \end{minipage}
    &
    \begin{minipage}{.23\textwidth}
      \includegraphics[width=\linewidth, height=32mm]{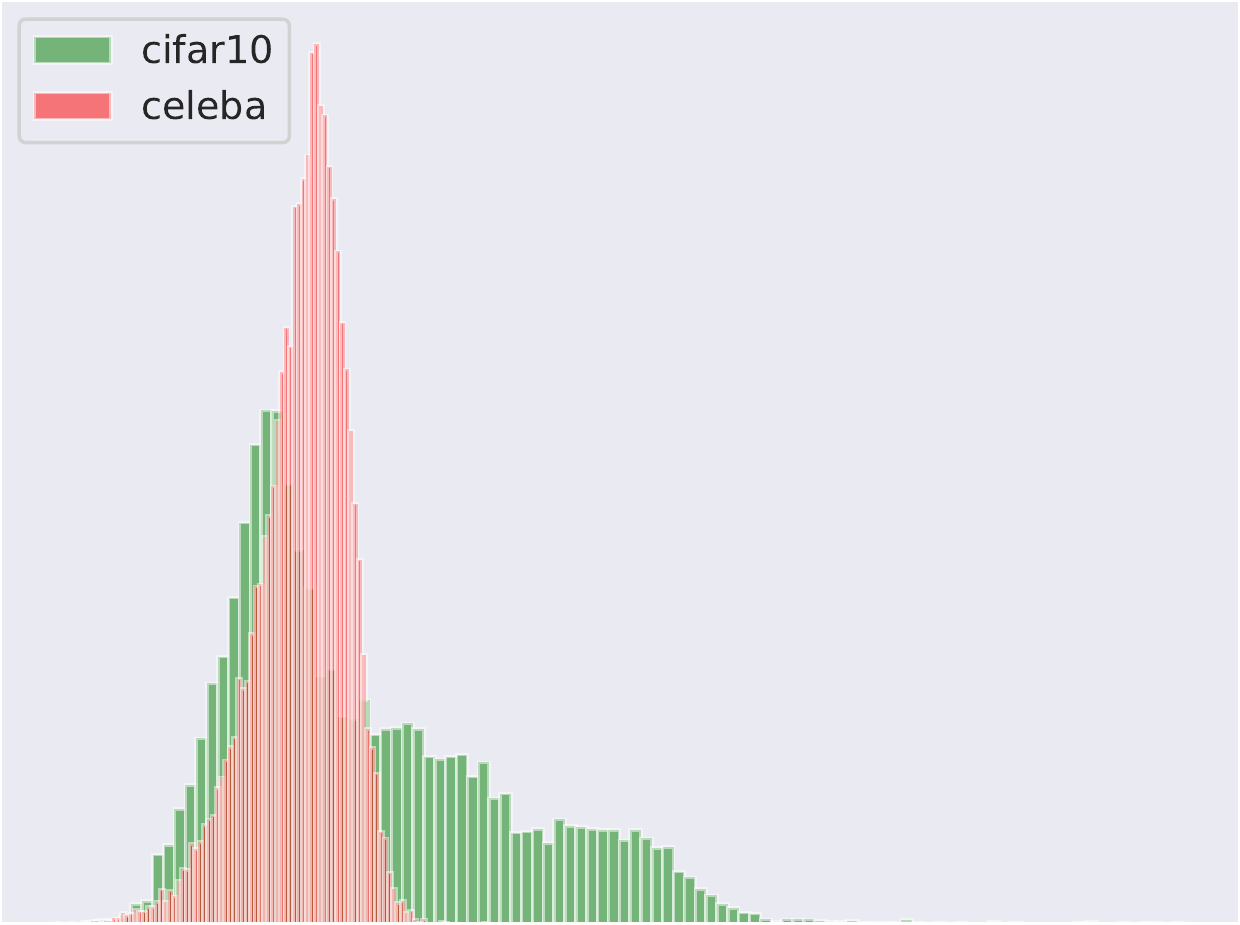}
    \end{minipage}
      \\ \hline
     SADA-JEM (K=10)
    &
    \begin{minipage}{.23\textwidth}
      \includegraphics[width=\linewidth, height=32mm]{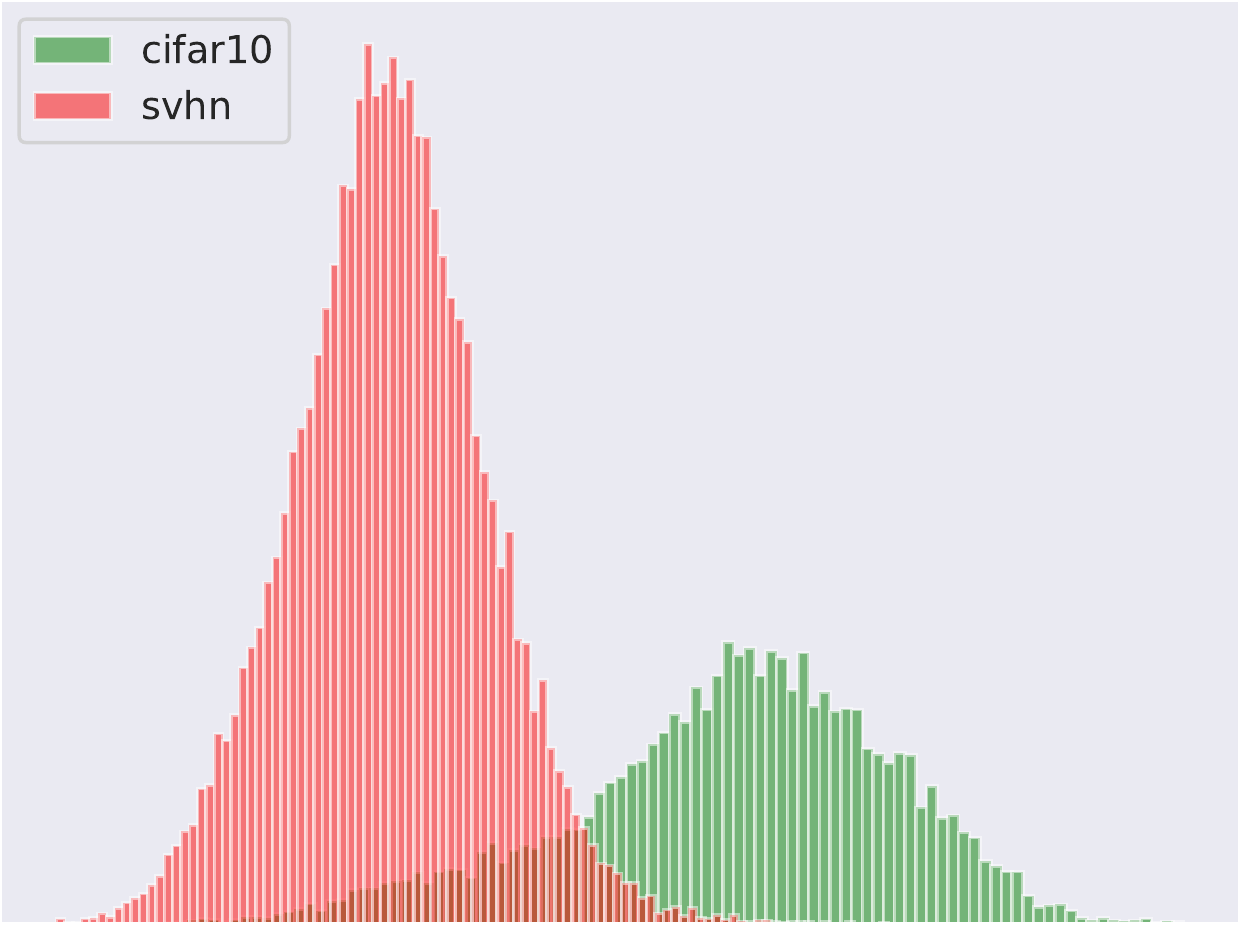}
    \end{minipage}
    &
    \begin{minipage}{.23\textwidth}
      \includegraphics[width=\linewidth, height=32mm]{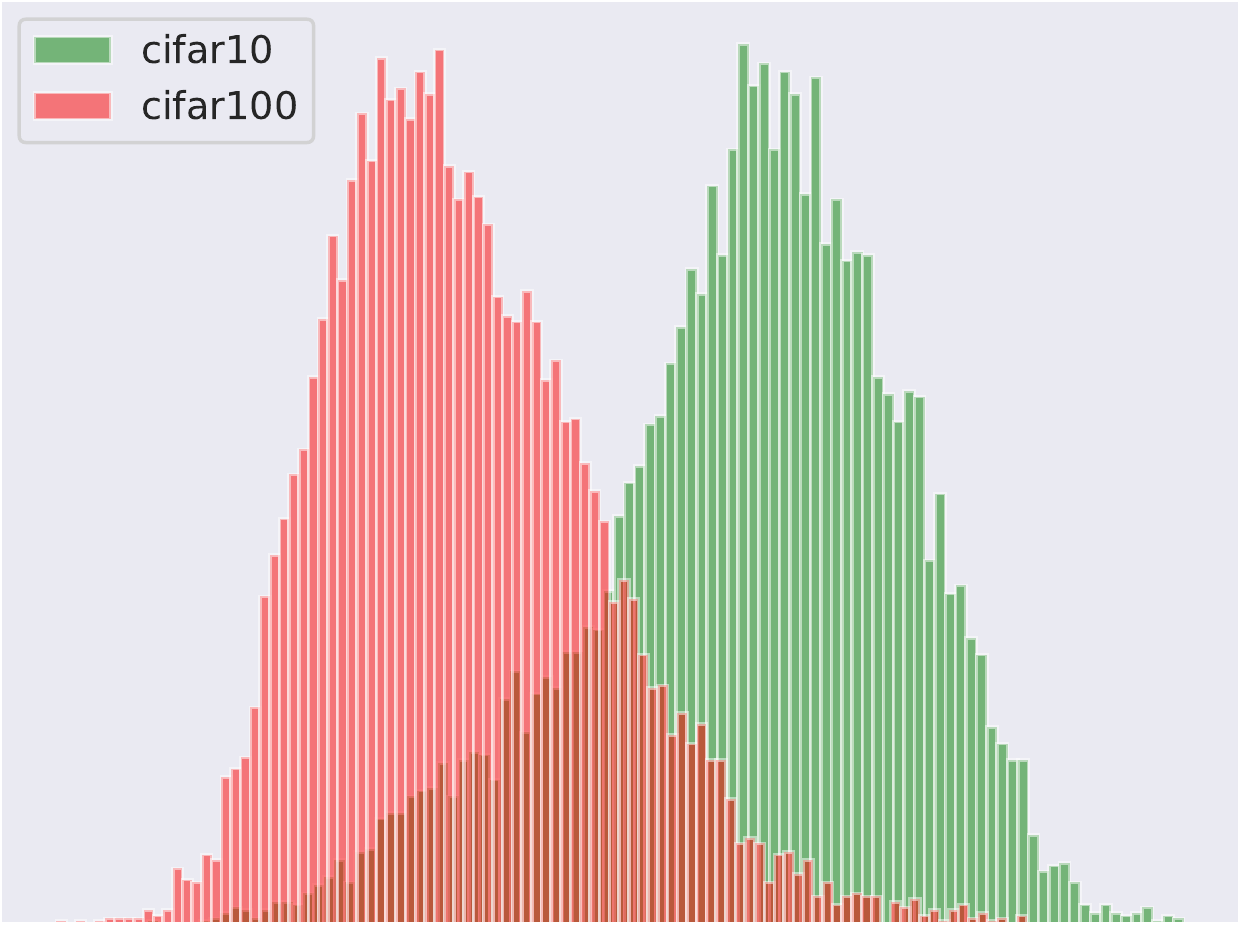}
    \end{minipage}
    &
    \begin{minipage}{.23\textwidth}
      \includegraphics[width=\linewidth, height=32mm]{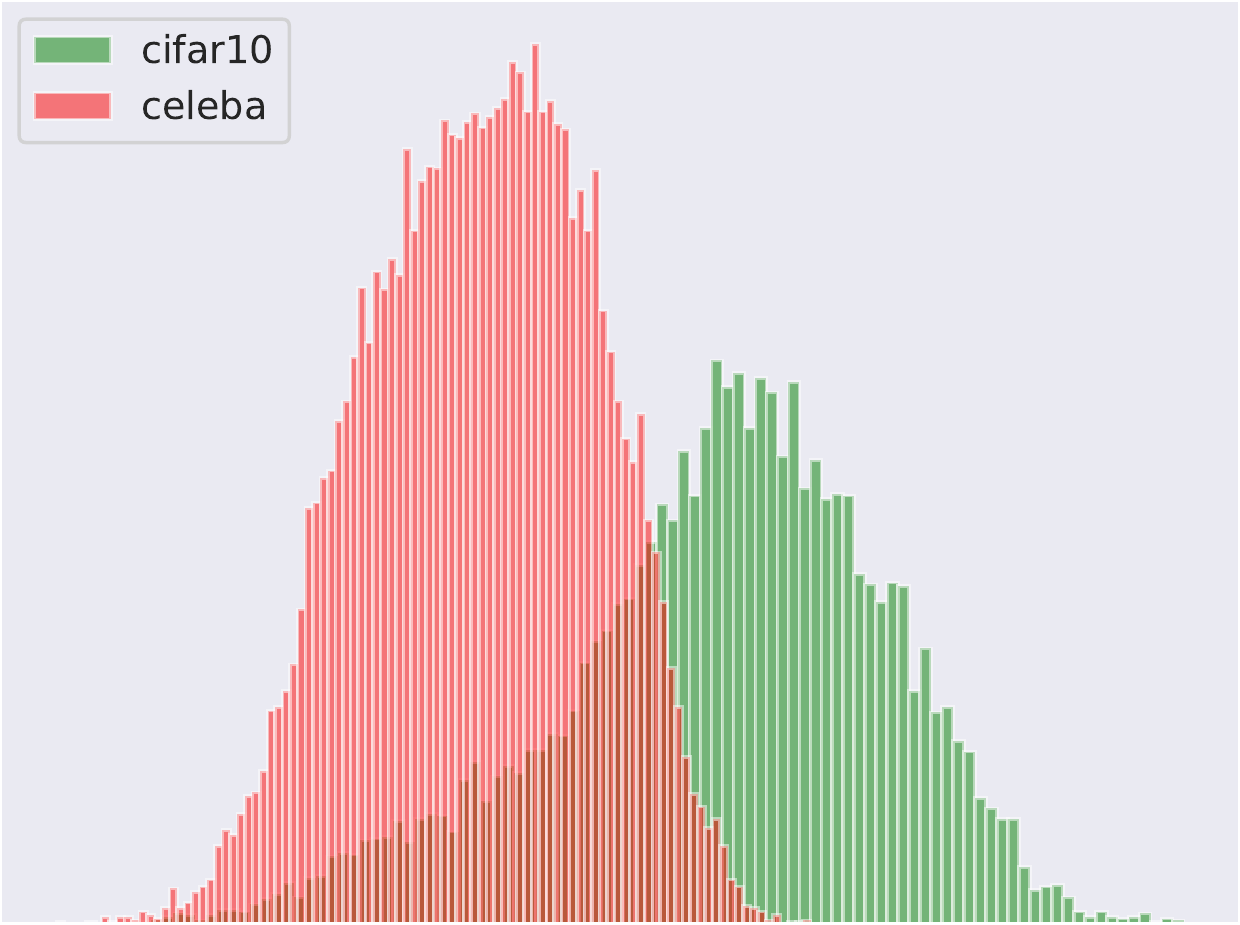}
    \end{minipage}
      \\ \hline
  \end{tabular}
  \caption{Histograms of $\log p_{\bs{\theta}}(\bs{x})$ for OOD detection. Green corresponds to in-distribution dataset, while red corresponds to OOD dataset.} 
  \label{table:logpx_hist}
\end{table*}

\subsection{Robustness}
DNNs are known to be vulnerable to adversarial examples~\cite{propertiesNN14, goodfellow2014explaining} in the form of tiny but sensitive perturbations to the inputs that trick the model to yield incorrect predictions. To mitigate this security threat posed by adversarial examples, a variety of defense algorithms have been proposed in the past few years to improve the robustness of deep networks~\cite{advexample15,dziugaite2016study,guo2017countering,akhtar2018defense,madry2018towards, chiang2020certified}. Existing works~\cite{jem,ebmdefense2021} have verified empirically that JEM is more robust than the softmax classifiers trained in standard procedures. Since SADA-JEM promotes the smoothness of energy landscape, it would be interesting to measure if SADA-JEM can also improve model robustness.

\begin{figure}[h]\vspace{-10pt}
  \centering
  \begin{subfigure}{0.48\linewidth}
    \includegraphics[width=1.1\columnwidth]{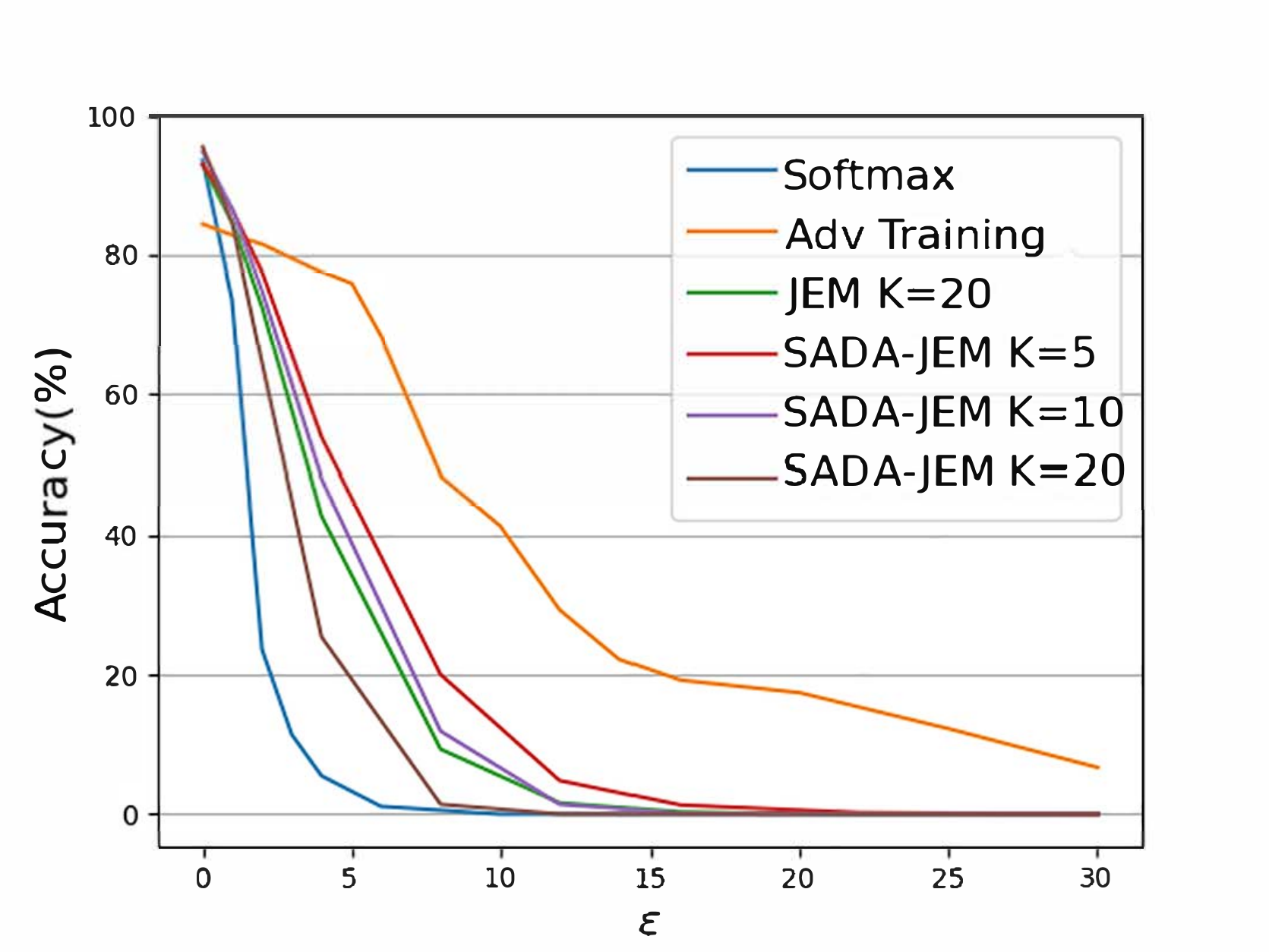}
    \caption{$L_\infty$ Robustness}
    \label{figure:l_inf_robust}
  \end{subfigure}
  \hfill
  \begin{subfigure}{0.48\linewidth}
    \includegraphics[width=1.1\columnwidth]{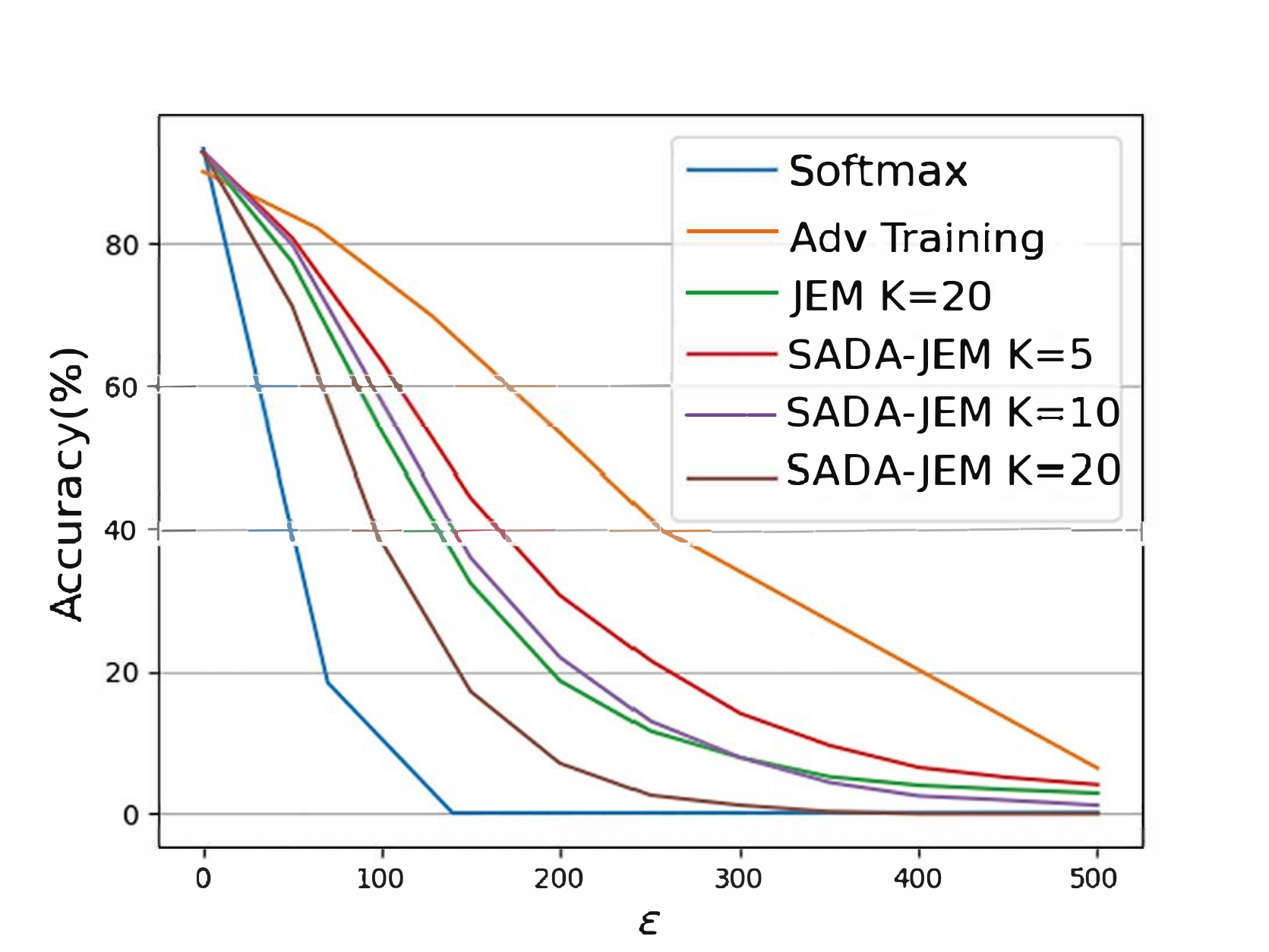}
    \caption{$L_2$ Robustness}
    \label{figure:l2_robust}
  \end{subfigure}\vspace{-5pt}
    \caption{Adversarial robustness under the PGD attacks.}
    \label{figure:robust_curve}\vspace{-10pt}
\end{figure}

The white-box PGD attack~\cite{madry2018towards} under an $L_\infty$ or $L_2$-norm constraint is the most common approach to evaluate the robustness of a classifier. However, Athalye et al.~\cite{Athalye2018} found that the defense methods using gradient obfuscation always report overrated robustness, and the defense can be overcome with minor adjustments to the standard PGD attacks. Therefore, to better evaluate the robustness of EBMs, Mitch Hill et al.~\cite{ebmdefense2021} proposed the Expectation-Over-Transformation (EOT) attack and Backward Pass Differentiable Approximation (BPDA) attack specifically for EBMs. We therefore employ these two attacks in our experiments  with the results reported in Figure~\ref{figure:robust_curve}.

As we can see, SADA-JEM achieves a similar robustness as JEM under the $L_\infty$ and $L_2$ PGD attacks, while both are more robust than the standard softmax classifiers. Moreover, we find that a larger $K$ undermines the robustness significantly, even though it can boost the accuracy on clean data. Similar observation has been reported by Yao et al.~\cite{jeat}, who found that EBM learns a smooth energy function around real data by increasing the energy of SGLD-sampled points; however, a larger $K$ can generate samples of lower energy which are closer to real data distribution, and thus leads to a sharper energy landscape around real data after optimizing on both real and generated samples. As such, the models trained with a larger $K$ are less robust than the ones with a smaller $K$. 
 In addition, JEM ($K=20$) diverges regularly, and it needs to restart the training by doubling $K$ (e.g., $K=40$), while SADA-JEM ($K=20$) is very stable. With a smaller $K$ SADA-JEM achieves even higher robustness than JEM ($K=20$).

\subsection{Ablation Study}

We study the impacts of SAM and data augmentation (DA) to the performance of SADA-JEM on image classification and generation in this section. The results on CIFAR10 are reported in Table~\ref{table:ablation}. It can be observed that SAM can improve the classification accuracy and generation quality of JEM/JEM++, while the improvements on classification accuracy are more pronounced. Secondly, by further excluding data augmentation $T$ from $p_{\bs{\theta}}(T(\bs{x}))$ of JEM++, which leads to SADA-JEM, the FID score is improved dramatically from 35.0 to 11.4. Prior works on EBMs~\cite{nijkamp2019learning,jem,jempp} include DA to their training pipelines to stabilize the training. However, DA introduces the artifacts to the training images, leading to foggy synthesized images. As a result, by excluding DA, SADA-JEM optimizes on $p_{\bs{\theta}}(\bs{x})$ and improves image generation quality significantly, while still being very stable due to the SAM optimizer. We further experiment replacing SAM in SADA-JEM with the energy $L_2$ regularization proposed in IGEBM~\cite{du2019implicit} to weakly regularize energy magnitudes of both positive and negative samples. We found that the $L_2$ regularization fails to improve the classification accuracy and degrades the training stability. 

We also study the impact of the noise radius $\rho$ to the performance of SADA-JEM in image classification and generation, with the results reported in Table~\ref{table:ablation_study_rho}. It can be observed that SAM with $\rho=0.2$ achieves an overall good performance in classification and image quality, and thus is chosen as default in all our experiments.

\begin{table}[ht!]
\caption{Ablation study of SADA-JEM. All the models are trained on CIFAR10 with $K=10$.}
\label{table:ablation}
\vspace{-15pt}
\begin{center}
\begin{threeparttable}
\begin{tabular}{l|cc}
\toprule
Ablation  & Acc\% $\uparrow$ & FID $\downarrow$  \\
\midrule
JEM                &  89.5  &  36.2   \\ 
JEM +SAM          &  90.1  &  35.0   \\
JEM++              &  93.5  &  37.1   \\
JEM++ +SAM        &  94.1  &  36.6   \\
JEM++ w/o DA       &  93.6  &  12.9   \\
JEM++ w/o DA +$L_2$*     &  93.4  &  -      \\  
SADA-JEM              &  \bf{96.0}  &  \bf{11.4}   \\
\bottomrule
\end{tabular}
\begin{tablenotes}
  \scriptsize\item * It fails to generate realistic images after 110 epochs.
\end{tablenotes}
\vspace{-10pt}
\end{threeparttable}
\vspace{-10pt}
\end{center}
\end{table}

\begin{table}[ht!]
\caption{Ablation study of SADA-JEM on $\rho$. All the models are trained on CIFAR10 with $K=10$.}
\label{table:ablation_study_rho}
\vspace{-15pt}
\begin{center}
\begin{threeparttable}\small
\begin{tabular}{l|cc}

\toprule
Ablation  & Acc \% $\uparrow$ & FID $\downarrow$  \\
\midrule
ASAM ($\rho=0.5$)            & 94.2    & 12.1    \\
ASAM ($\rho=1$)              & 94.5    & 11.9    \\
ASAM ($\rho=2$)              & 94.8    & 11.7    \\
ASAM ($\rho=4$)              & 95.3    & 11.5    \\
ASAM ($\rho=8$)              & \multicolumn{2}{c}{Diverged after 2nd epoch}    \\
\midrule
SAM ($\rho=0.05$)           &  94.8     &   \bf{10.9}      \\
SAM ($\rho=0.1$)            &  95.5     &   11.4      \\
SAM ($\rho=0.2$)            &  \bf{96.0}     &   11.4     \\
SAM ($\rho=0.4$)            &  95.1     &   14.1      \\
SAM ($\rho=0.8$)            &  91.9     &   19.5      \\
\bottomrule
\end{tabular}
\end{threeparttable}
\vspace{-10pt}
\end{center}
\end{table}

\section{Limitations}\label{sec:limit}

It is challenging to train SGLD-based EBMs, including IGEBM, JEM, JEM++ and SADA-JEM, on complex high-dimensional data. IGEBM, JEM and many prior works have investigated methods to stabilize the training of EBM, but they require an extremely expensive SGLD sampling with a large $K$. Our SADA-JEM can stabilize the training on CIFAR10 and CIFAR100 with a small $K$ (e.g., $K\!=\!5$). However, when the image resolution scales up (e.g., from 32x32 to 224x224), SADA-JEM has to increase $K$ accordingly to improve image generation quality. Hence, the trade-off between generation quality and computational complexity still limits the application of SADA-JEM to large-scale benchmarks, including ImageNet~\cite{imagenet09}. 

Besides, the computation bottleneck of SADA-JEM is not SAM as SAM is only used to optimize model parameters $\bs{\theta}$ (the outer maximization
in Eq.~\ref{eq:sajem_obj}). Instead, the $K$ SGLD sampling steps (typically $K\!=\!10$) is the most expensive operation (the inner minimization
in Eq.~\ref{eq:sajem_obj}). SAM doubles the cost of $\bs{\theta}$ optimization, which is insignificant compared to $K$ SGLD steps.
Overall, the training speed of SADA-JEM is comparable to JEM/JEM++. Therefore, a more efficient sampling method is required to scale up SADA-JEM to large-scale applications. 


\section{Conclusion}
We propose SADA-JEM to bridge the classification accuracy gap and the generation quality gap of JEM. By incorporating the framework of SAM to JEM and excluding the undesirable data augmentation from the training pipeline of JEM, SADA-JEM promotes the energy landscape smoothness and hence the generalization of trained models. Our experiments verify the effectiveness of these techniques on multiple benchmarks and demonstrate the state-of-the-art results in most of the tasks of image classification, generation, uncertainty calibration, OOD detection and adversarial robustness. As for the future work, we are interested in improving the scalability of EBMs to large-scale benchmarks, such as ImageNet and NLP tasks.

\section{Acknowledgement}
We would like to thank the anonymous reviewers for their comments and suggestions, which
helped improve the quality of this paper. We would also gratefully acknowledge the support
of Cisco Systems, Inc. for its university research fund to this research.

{\small
\bibliographystyle{ieee_fullname}
\bibliography{ml}
}

\appendix



\section{Experimental Details}\label{app:exp}

To have a fair comparison, we largely follow the settings of JEM~\cite{jem} and JEM++~\cite{jempp}, and train our models based on the Wide-ResNet 28x10 architecture~\cite{wideresnet16} for 200 epochs. We use SGD for CIFAR10 and CIFAR100 with an initial learning rate of 0.1 and 0.01, respectively, and decay the learning rate by 0.2 at epoch [60, 120, 180] for most cases. Apart from this, we find that the cosine learning rate scheduler can be adopted for SADA-JEM, which achieves much better accuracy and FID on CIFAR10.\footnote{This is because the combination of SAM and single branched DA improves the training stability
significantly. As a result, the cosine learning rate decay can be adopted to improve the overall performance. JEM, JEM++ and other SADA-JEM ablation configurations are less stable to enable the cosine learning rate decay.} The hyper-parameters used in our experiments are listed in Table~\ref{table:hyperparameters}.

\begin{table}[ht!]
\caption{Hyper-parameters of SADA-JEM for CIFAR10 and CIFAR100.}
\label{table:hyperparameters}\vspace{-15pt}
\begin{center}
\begin{threeparttable}
\begin{tabular}{l|cc}
\toprule
Variable      & Value \\
\midrule
Number of SGLD steps $K$             & 5, 10, 20   \\
Buffer size $|\mathbb{B}|$           & 10,000      \\
Reinitialization freq. $\gamma$      & 5\%         \\
SGLD step-size $\alpha$              &  1          \\
SGLD noise $\sigma$                  &  0          \\
SAM noise radius $\rho$               &  0.2        \\
\bottomrule
\end{tabular}
\end{threeparttable}
\end{center}\vspace{-10pt}
\end{table}

\section{Visualizing Generated Images}
\begin{figure*}
  \centering
  \begin{subfigure}{0.31\linewidth}
    \includegraphics[width=1\columnwidth]{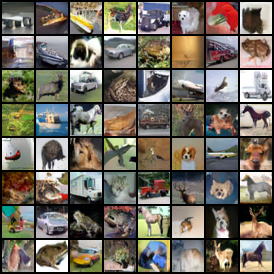}
    \caption{SADA-JEM}
    \label{figure:sajem_vs}
  \end{subfigure}
  \hfill
\begin{subfigure}{0.31\linewidth}
    \includegraphics[width=1\columnwidth]{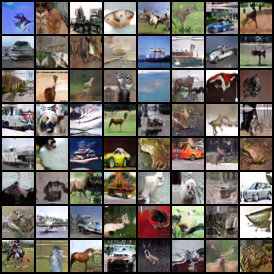}
    \caption{VERA}
    \label{figure:vera_vs}
  \end{subfigure}
  \hfill
  \begin{subfigure}{0.31\linewidth}
    \includegraphics[width=1\columnwidth]{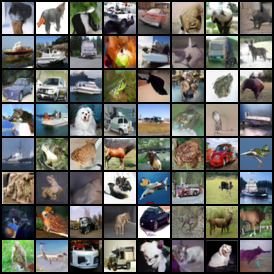}
    \caption{DiffuRecov}
    \label{figure:diffrec_vs}
  \end{subfigure}
  \hfill

\caption{Generated images from SADA-JEM, VERA, and DiffuRecov.}
\label{figure:sajem_vera_dif}
\end{figure*}

Table 1 in the main text reports the quantitative performance comparison of different stand-alone generative models and hybrid models. Here in Figure~\ref{figure:sajem_vera_dif} we provide a qualitative comparison of generated images from (a) SADA-JEM, (b) VERA~\cite{nomcmc}, and (c) DiffuRecov~\cite{diffusionRecovery}. As we can see, the perceived image qualities of them are comparable even though DiffuRecov has a much better FID score than that of VERA (9.58 vs. 30.5), indicating that visualizing generated images is less effective to evaluate image quality.

\begin{figure*}
  \centering
  \begin{subfigure}{0.31\linewidth}
    \includegraphics[width=1\columnwidth]{figures/wrn2810_e_3d.pdf}
    \caption{Classifier}
    \label{figure:e_landscape_cls2}
  \end{subfigure}
  \hfill
\begin{subfigure}{0.31\linewidth}
    \includegraphics[width=1\columnwidth]{figures/jem_89.57_e_rev_3d.pdf}
    \caption{JEM}
    \label{figure:e_landscape_official_jem2}
  \end{subfigure}
  \hfill
  \begin{subfigure}{0.31\linewidth}
    \includegraphics[width=1\columnwidth]{figures/jemsam_energy_3d.pdf}
    \caption{JEM+SAM}
    \label{figure:e_landscape_jemsam2}
  \end{subfigure}
  \hfill
  
  \begin{subfigure}{0.31\linewidth}
    \includegraphics[width=1\columnwidth]{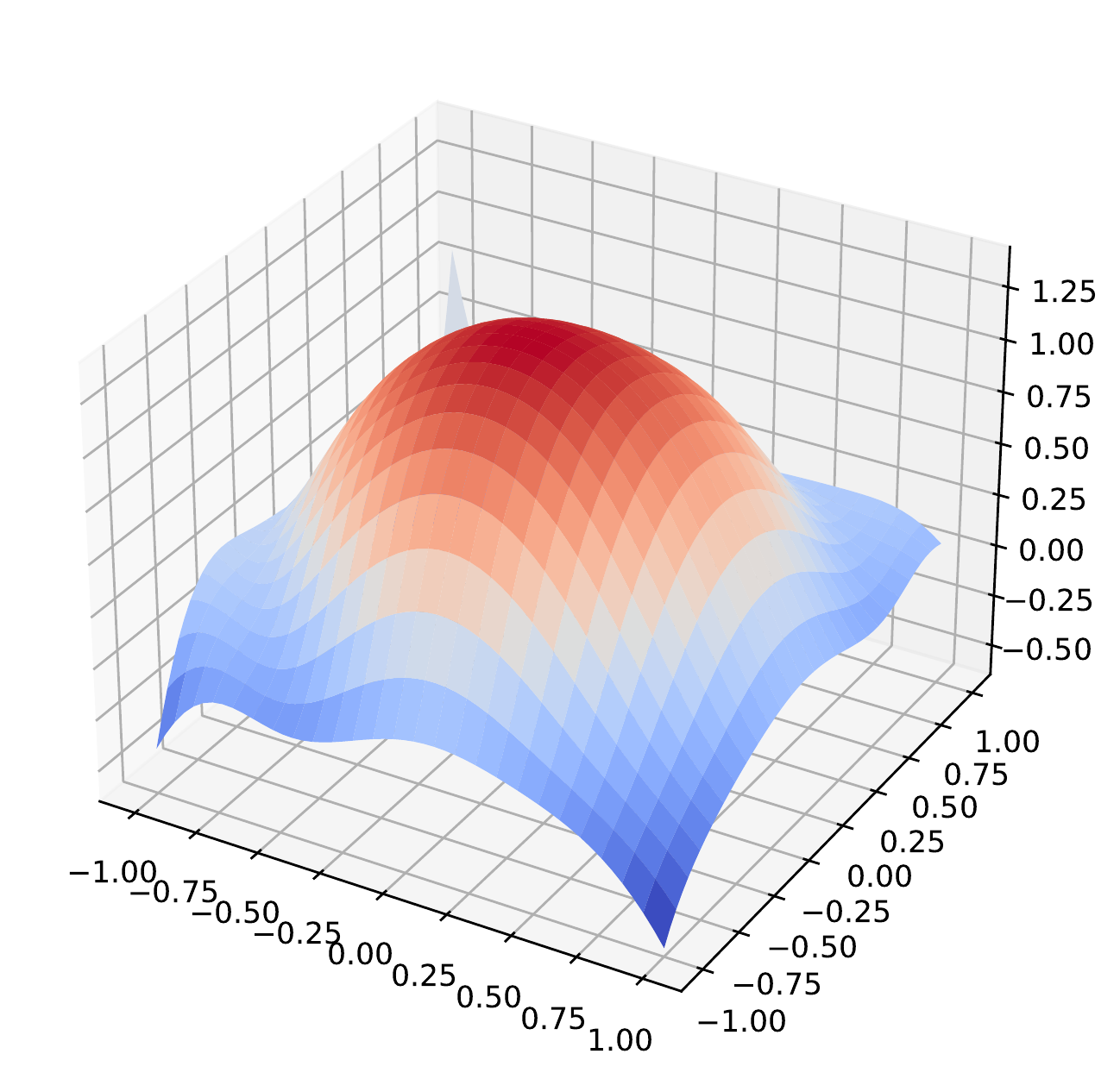}
    \caption{SADA-JEM (K=5)}
    \label{figure:e_landscape_sajem5_app}
  \end{subfigure}
  \hfill
\begin{subfigure}{0.31\linewidth}
    \includegraphics[width=1\columnwidth]{figures/sajem10_energy_3d.pdf}
    \caption{SADA-JEM (K=10)}
    \label{figure:e_landscape_sajem10_app}
  \end{subfigure}
  \hfill
  \begin{subfigure}{0.31\linewidth}
    \includegraphics[width=1\columnwidth]{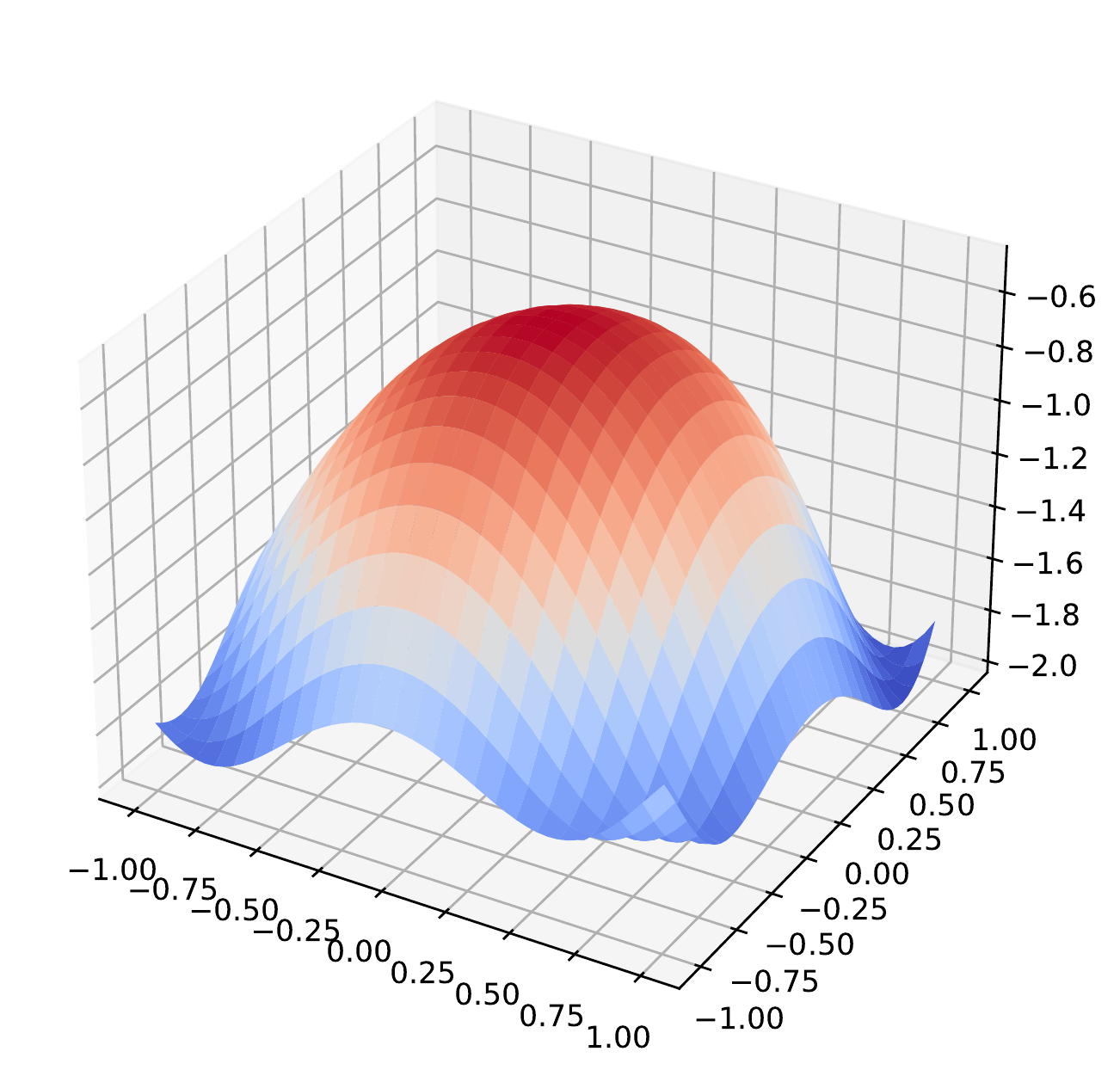}
    \caption{SADA-JEM (K=20)}
    \label{figure:e_landscape_sajem20_app}
  \end{subfigure}
  \hfill
\caption{Energy landscapes of different models trained on CIFAR10. Please note the different scales of the y-axes.}
\label{figure:energy_landscape_app}
\end{figure*}

\section{Energy Landscapes}\vspace{-5pt}

Figure~\ref{figure:energy_landscape_app} illustrates the energy landscapes of different models trained on CIFAR10. The energy landscape is generated by visualizing $E(\bs{\theta})=\sum_{\bs{x}\in{\bs{X}}}E_{\bs{\theta}}(\bs{x})$ with the technique introduced in~\cite{losslandscape}, where $X$ is a 10\% random samples from CIFAR10 training data. As we can see, SADA-JEM's energy landscapes are much smoother than those of the competing methods (see different scales of the y-axes).

\section{Out-of-Distribution Detection}\vspace{-5pt}

Table~\ref{table:logpx_hist_app} reports the OOD detection performances of different models and SADA-JEM with different $K$s, where the input density $\log p_{\bs{\theta}}(\bs{x})$ is used as $s_{\bs{\theta}}(\bs{x})$ for OOD detection on CIFAR10.

\begin{table*}[ht!]
  \centering
  \begin{tabular}{ | c | m{4.2cm} | m{4.2cm} | m{4.2cm} | }
    \hline
    JEM
    &
    \begin{minipage}{.25\textwidth}
      \includegraphics[width=\linewidth, height=35mm]{figures/jem_svhn_logp.pdf}
    \end{minipage}
    &
    \begin{minipage}{.25\textwidth}
      \includegraphics[width=\linewidth, height=35mm]{figures/jem_CIFAR100_logp.pdf}
    \end{minipage}
    &
    \begin{minipage}{.25\textwidth}
      \includegraphics[width=\linewidth, height=35mm]{figures/jem_celeba_logp.pdf}
    \end{minipage}
      \\ \hline
    JEM++ (M=10)
    &
    \begin{minipage}{.25\textwidth}
      \includegraphics[width=\linewidth, height=35mm]{figures/jempp10_svhn_logp.pdf}
    \end{minipage}
    &
    \begin{minipage}{.25\textwidth}
      \includegraphics[width=\linewidth, height=35mm]{figures/jempp10_cifar100_logp.pdf}
    \end{minipage}
    &
    \begin{minipage}{.25\textwidth}
      \includegraphics[width=\linewidth, height=35mm]{figures/jempp10_celeba_logp.pdf}
    \end{minipage}
      \\ \hline
     SADA-JEM (K=5)
    &
    \begin{minipage}{.25\textwidth}
      \includegraphics[width=\linewidth, height=35mm]{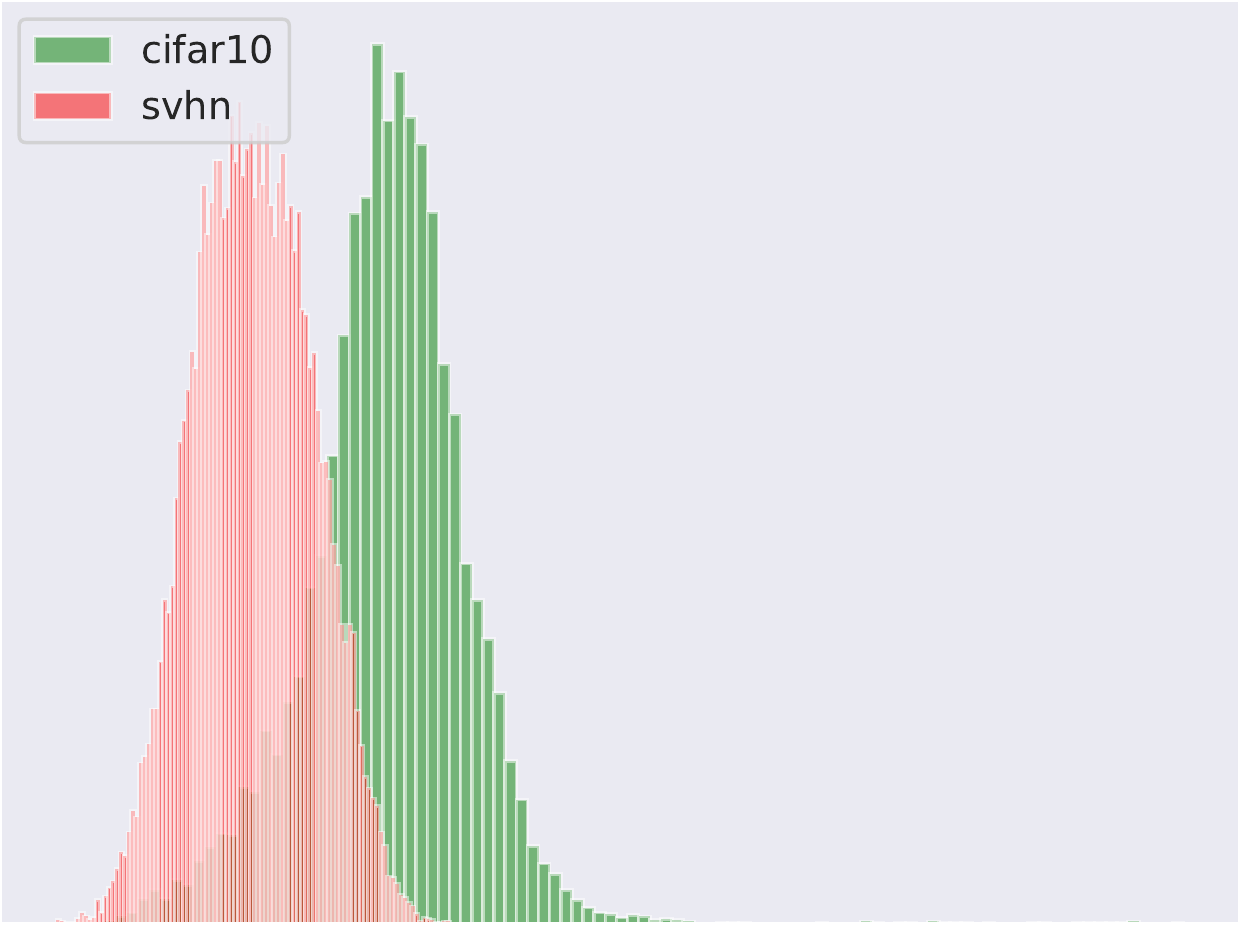}
    \end{minipage}
    &
    \begin{minipage}{.25\textwidth}
      \includegraphics[width=\linewidth, height=35mm]{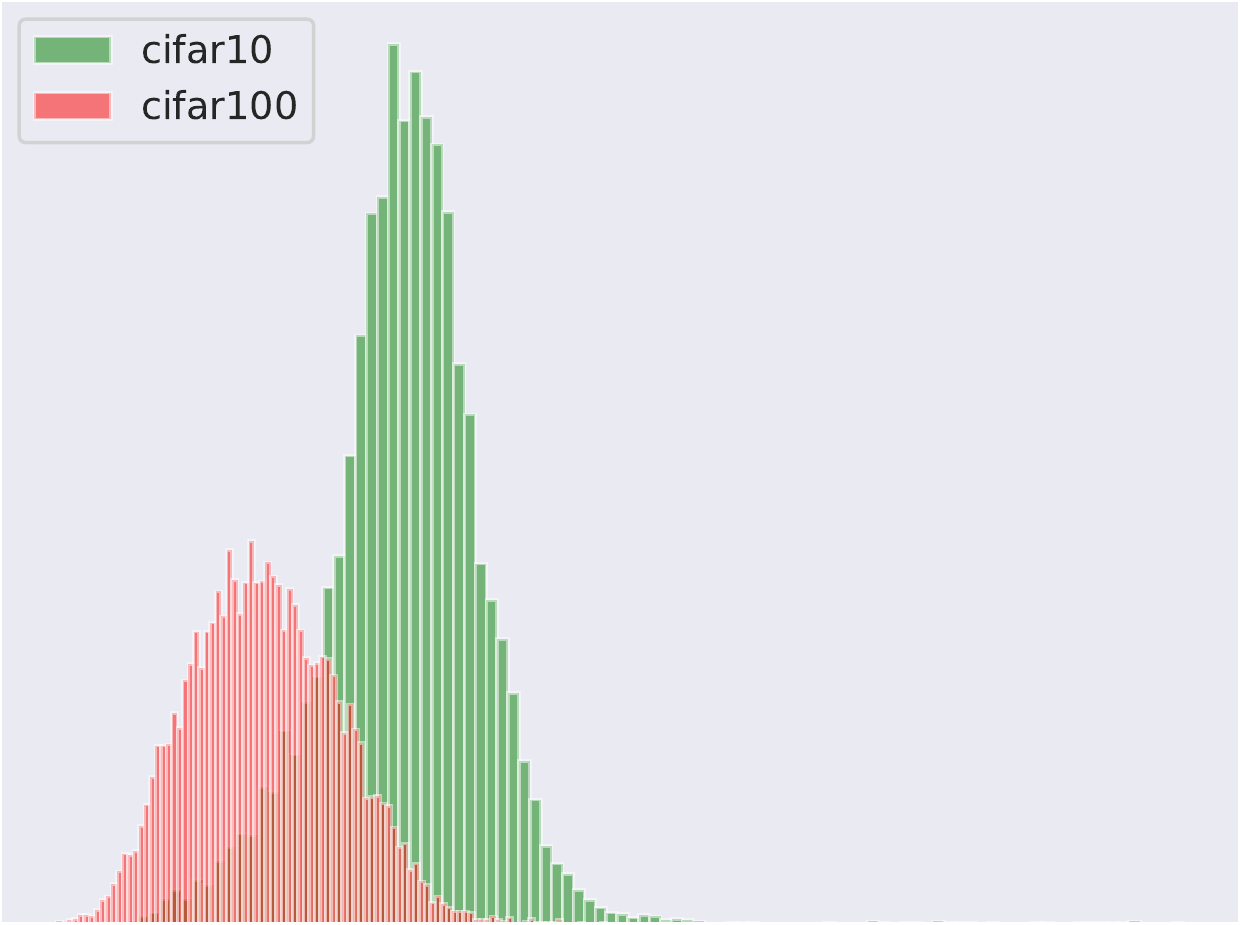}
    \end{minipage}
    &
    \begin{minipage}{.25\textwidth}
      \includegraphics[width=\linewidth, height=35mm]{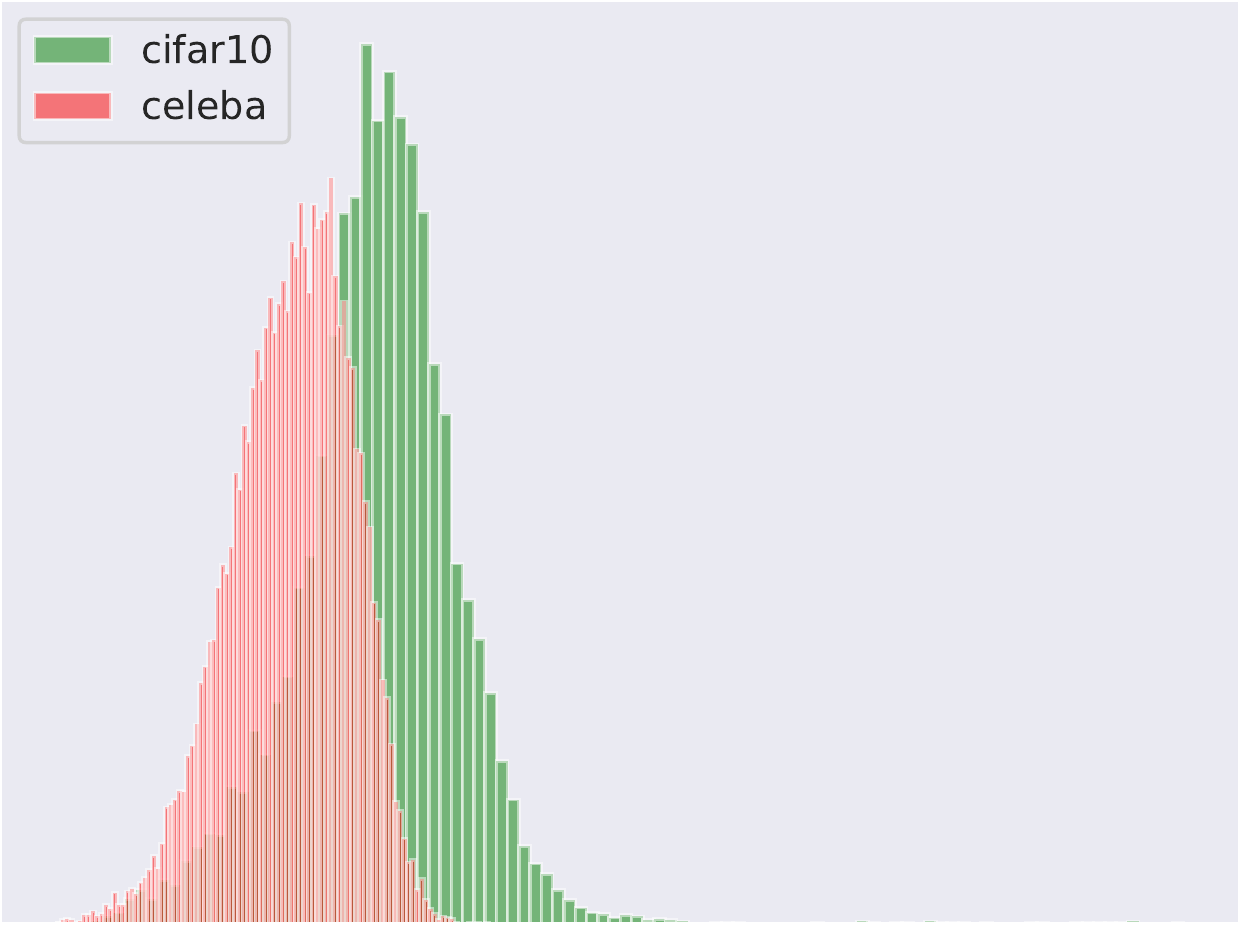}
    \end{minipage}
      \\ \hline
     SADA-JEM (K=10)
    &
    \begin{minipage}{.25\textwidth}
      \includegraphics[width=\linewidth, height=35mm]{figures/sajem_svhn_logp.pdf}
    \end{minipage}
    &
    \begin{minipage}{.25\textwidth}
      \includegraphics[width=\linewidth, height=35mm]{figures/sajem_CIFAR100_logp.pdf}
    \end{minipage}
    &
    \begin{minipage}{.25\textwidth}
      \includegraphics[width=\linewidth, height=35mm]{figures/sajem_celeba_logp.pdf}
    \end{minipage}
      \\ \hline
     SADA-JEM (K=20)
    &
    \begin{minipage}{.25\textwidth}
      \includegraphics[width=\linewidth, height=35mm]{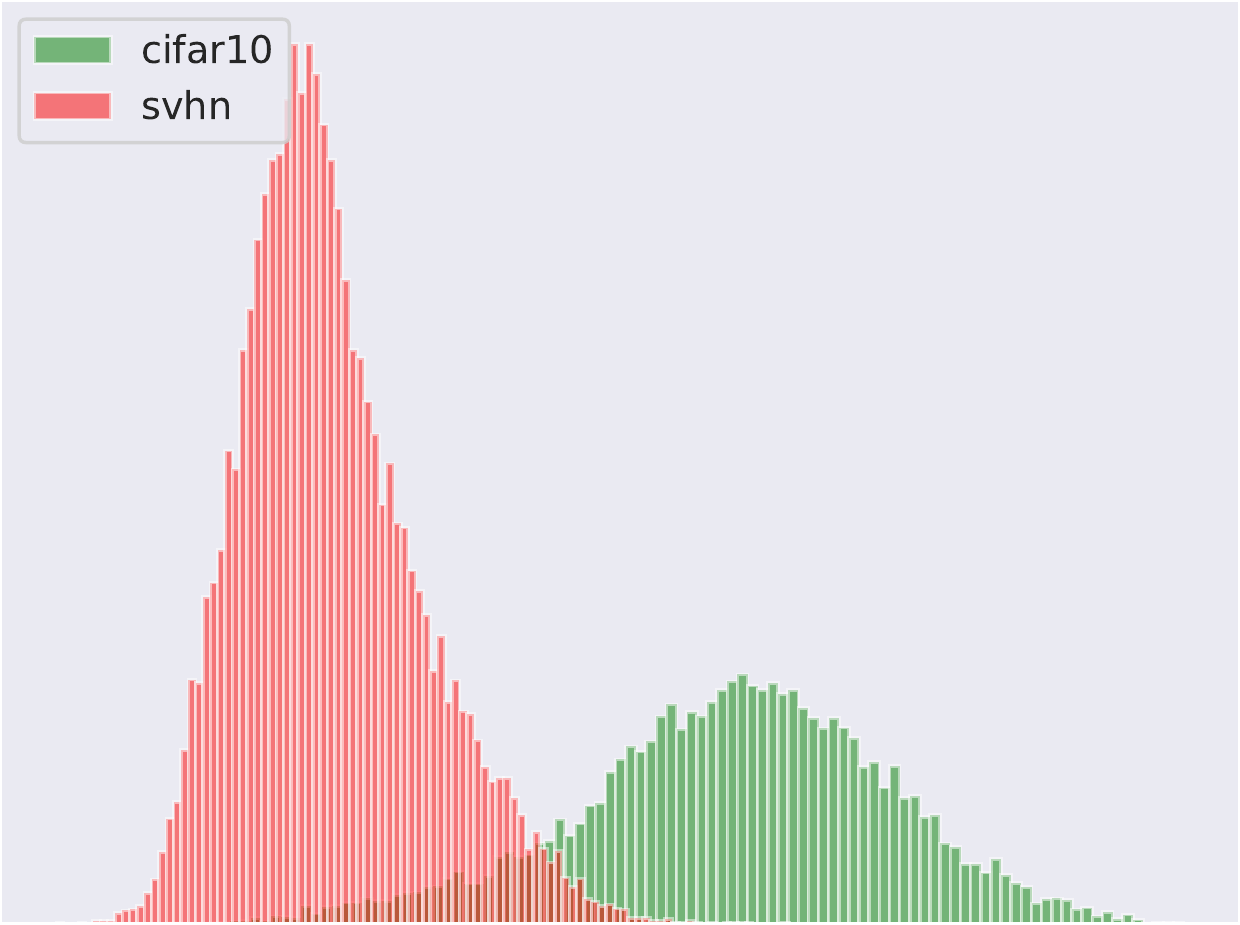}
    \end{minipage}
    &
    \begin{minipage}{.25\textwidth}
      \includegraphics[width=\linewidth, height=35mm]{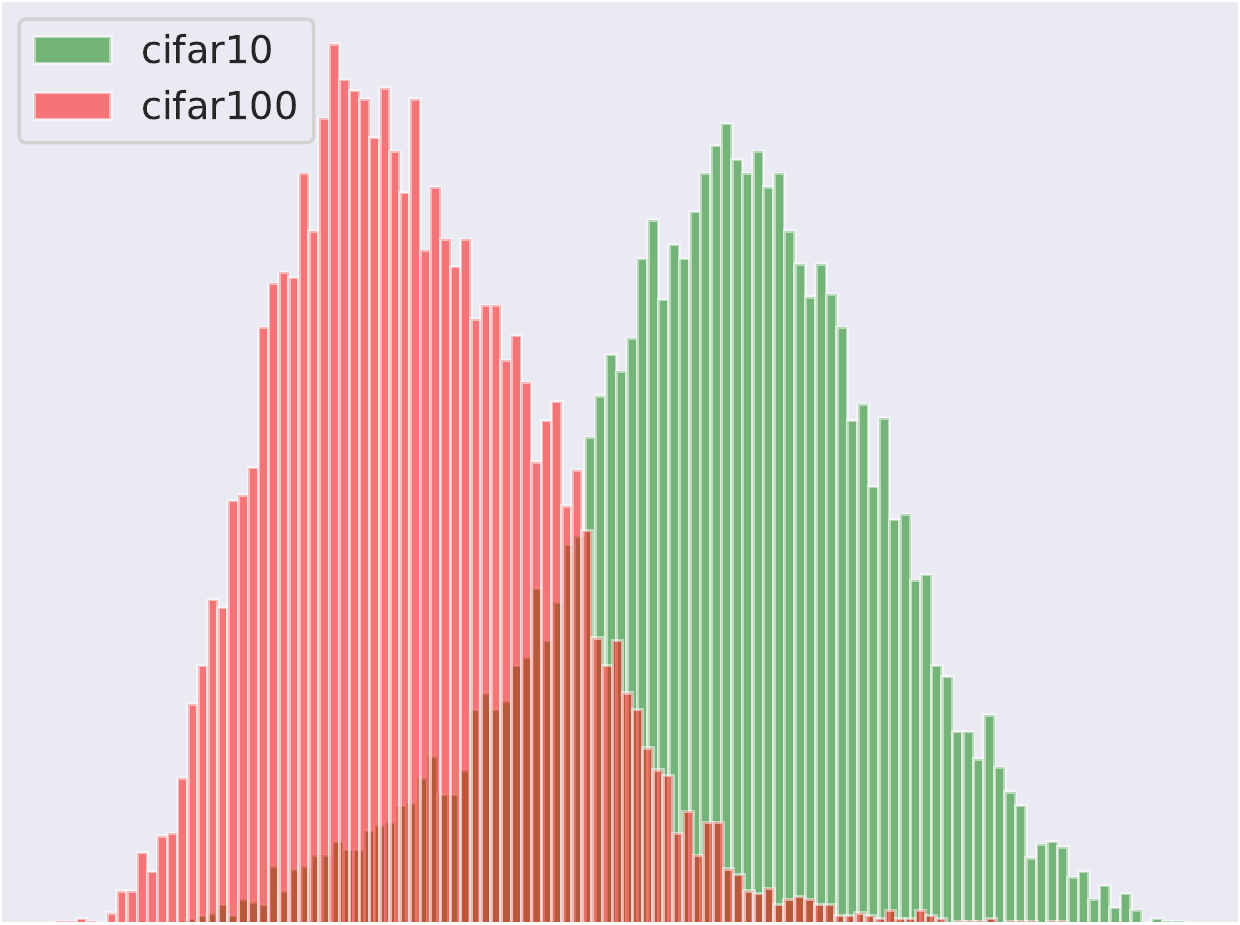}
    \end{minipage}
    &
    \begin{minipage}{.25\textwidth}
      \includegraphics[width=\linewidth, height=35mm]{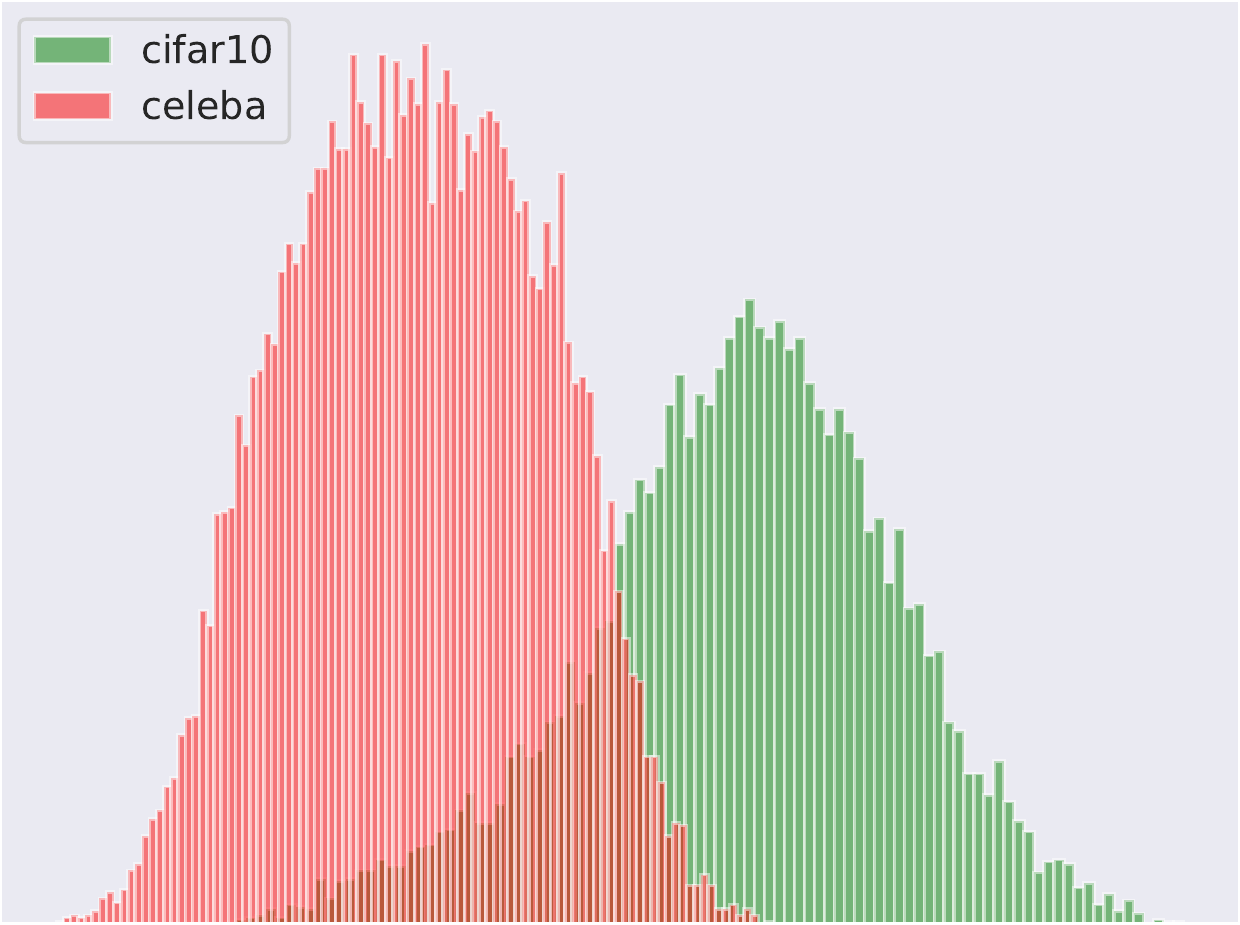}
    \end{minipage}
      \\ \hline
  \end{tabular}
  \caption{Histograms of $\log p_{\bs{\theta}}(\bs{x})$ for OOD detection. Green corresponds to in-distribution dataset, while red corresponds to OOD dataset.} 
  \label{table:logpx_hist_app}
\end{table*}

\vspace{-5pt}
\section{Additional Generated Samples}\label{app:samples}\vspace{-5pt}

\begin{figure*}
  \centering
  \begin{subfigure}{0.23\linewidth}
    \includegraphics[width=1\columnwidth]{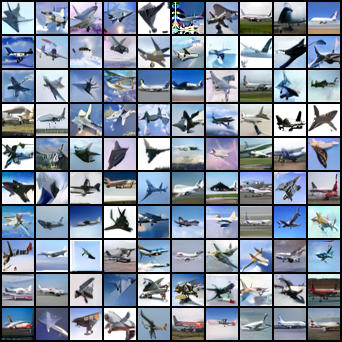}
    \caption{Samples with highest $p(\bs{x}$)}
  \end{subfigure}
  \hfill
  \begin{subfigure}{0.23\linewidth}
    \includegraphics[width=1\columnwidth]{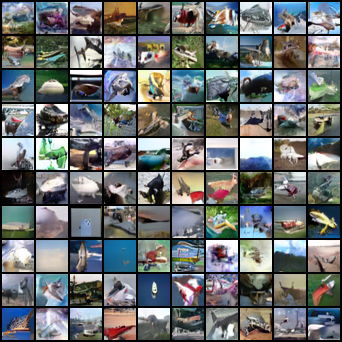}
    \caption{Samples with lowest $p(\bs{x}$)}
  \end{subfigure}
  \hfill
  \begin{subfigure}{0.23\linewidth}
    \includegraphics[width=1\columnwidth]{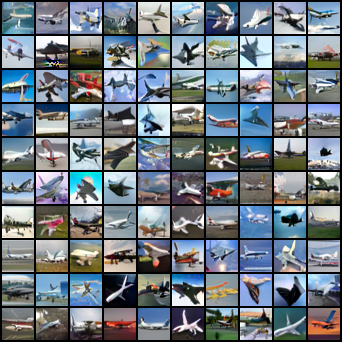}
    \caption{Samples with highest $p(y|\bs{x}$)}
  \end{subfigure}
  \hfill
  \begin{subfigure}{0.23\linewidth}
    \includegraphics[width=1\columnwidth]{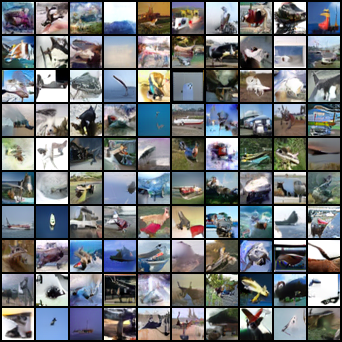}
    \caption{Samples with lowest $p(y|\bs{x}$)}
  \end{subfigure}
  \hfill
\caption{SADA-JEM generated class-conditional samples of \textbf{Plane}.}
\label{figure:sajem_app_class_0}
\end{figure*}

\begin{figure*}
  \centering
  \begin{subfigure}{0.23\linewidth}
    \includegraphics[width=1\columnwidth]{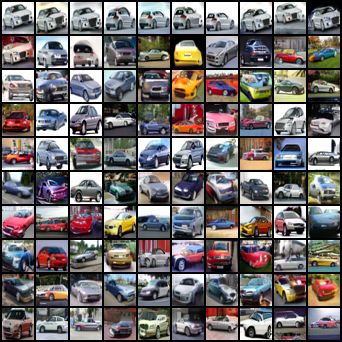}
    \caption{Samples with highest $p(\bs{x}$)}
  \end{subfigure}
  \hfill
  \begin{subfigure}{0.23\linewidth}
    \includegraphics[width=1\columnwidth]{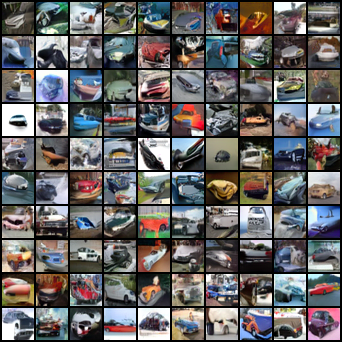}
    \caption{Samples with lowest $p(\bs{x}$)}
  \end{subfigure}
  \hfill
  \begin{subfigure}{0.23\linewidth}
    \includegraphics[width=1\columnwidth]{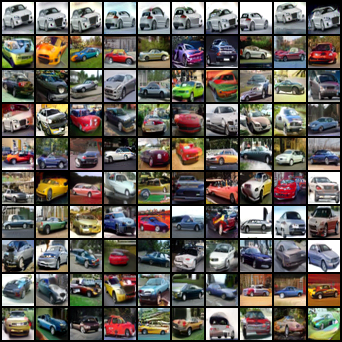}
    \caption{Samples with highest $p(y|\bs{x}$)}
  \end{subfigure}
  \hfill
  \begin{subfigure}{0.23\linewidth}
    \includegraphics[width=1\columnwidth]{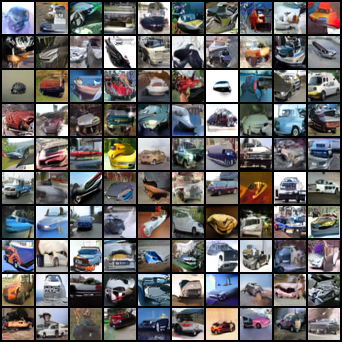}
    \caption{Samples with lowest $p(y|\bs{x}$)}
  \end{subfigure}
  \hfill
\caption{SADA-JEM generated class-conditional samples of \textbf{Car}.}
\label{figure:sajem_app_class_1}
\end{figure*}

\begin{figure*}
  \centering
  \begin{subfigure}{0.23\linewidth}
    \includegraphics[width=1\columnwidth]{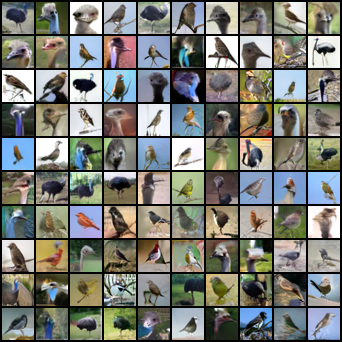}
    \caption{Samples with highest $p(\bs{x}$)}
  \end{subfigure}
  \hfill
  \begin{subfigure}{0.23\linewidth}
    \includegraphics[width=1\columnwidth]{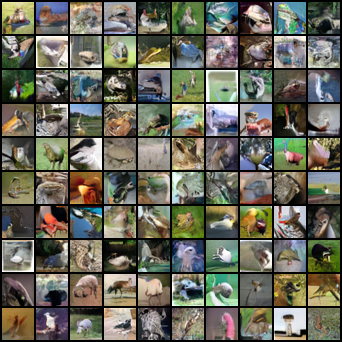}
    \caption{Samples with lowest $p(\bs{x}$)}
  \end{subfigure}
  \hfill
  \begin{subfigure}{0.23\linewidth}
    \includegraphics[width=1\columnwidth]{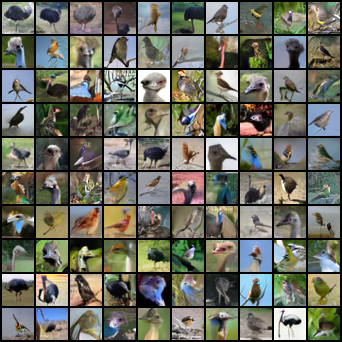}
    \caption{Samples with highest $p(y|\bs{x}$)}
  \end{subfigure}
  \hfill
  \begin{subfigure}{0.23\linewidth}
    \includegraphics[width=1\columnwidth]{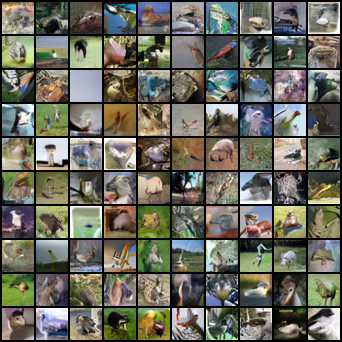}
    \caption{Samples with lowest $p(y|\bs{x}$)}
  \end{subfigure}
  \hfill
\caption{SADA-JEM generated class-conditional samples of \textbf{Bird}.}
\label{figure:sajem_app_class_2}
\end{figure*}

\begin{figure*}
  \centering
  \begin{subfigure}{0.23\linewidth}
    \includegraphics[width=1\columnwidth]{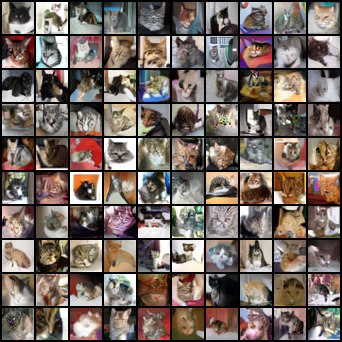}
    \caption{Samples with highest $p(\bs{x}$)}
  \end{subfigure}
  \hfill
  \begin{subfigure}{0.23\linewidth}
    \includegraphics[width=1\columnwidth]{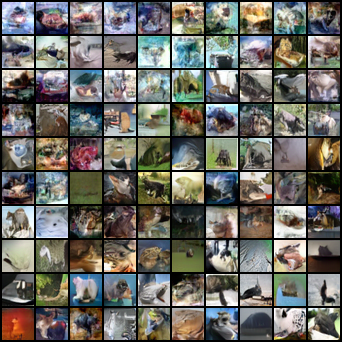}
    \caption{Samples with lowest $p(\bs{x}$)}
  \end{subfigure}
  \hfill
  \begin{subfigure}{0.23\linewidth}
    \includegraphics[width=1\columnwidth]{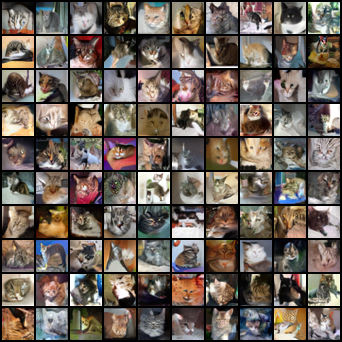}
    \caption{Samples with highest $p(y|\bs{x}$)}
  \end{subfigure}
  \hfill
  \begin{subfigure}{0.23\linewidth}
    \includegraphics[width=1\columnwidth]{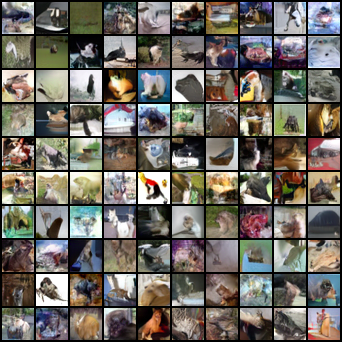}
    \caption{Samples with lowest $p(y|\bs{x}$)}
  \end{subfigure}
  \hfill
\caption{SADA-JEM generated class-conditional samples of \textbf{Cat}.}
\label{figure:sajem_app_class_3}
\end{figure*}

\begin{figure*}
  \centering
  \begin{subfigure}{0.23\linewidth}
    \includegraphics[width=1\columnwidth]{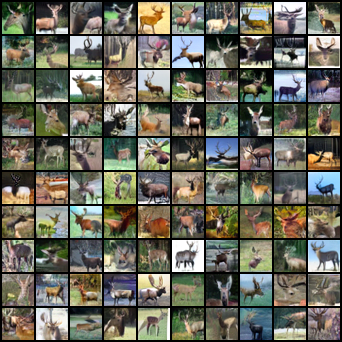}
    \caption{Samples with highest $p(\bs{x}$)}
  \end{subfigure}
  \hfill
  \begin{subfigure}{0.23\linewidth}
    \includegraphics[width=1\columnwidth]{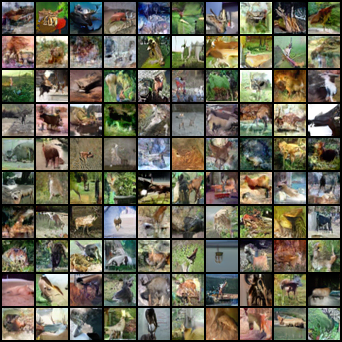}
    \caption{Samples with lowest $p(\bs{x}$)}
  \end{subfigure}
  \hfill
  \begin{subfigure}{0.23\linewidth}
    \includegraphics[width=1\columnwidth]{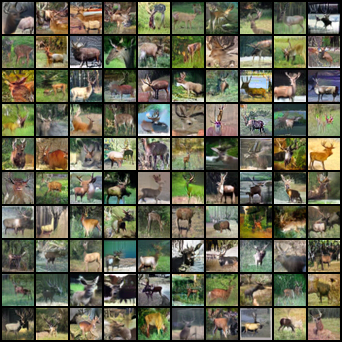}
    \caption{Samples with highest $p(y|\bs{x}$)}
  \end{subfigure}
  \hfill
  \begin{subfigure}{0.23\linewidth}
    \includegraphics[width=1\columnwidth]{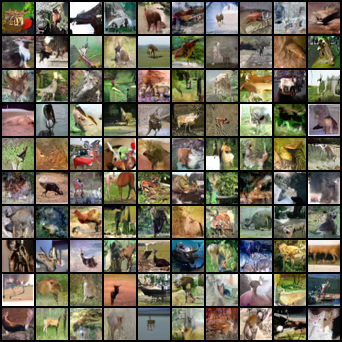}
    \caption{Samples with lowest $p(y|\bs{x}$)}
  \end{subfigure}
  \hfill
\caption{SADA-JEM generated class-conditional samples of \textbf{Deer}.}
\label{figure:sajem_app_class_4}
\end{figure*}

\begin{figure*}
  \centering
  \begin{subfigure}{0.23\linewidth}
    \includegraphics[width=1\columnwidth]{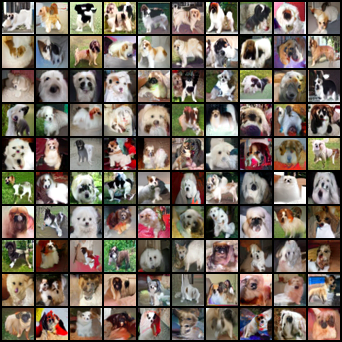}
    \caption{Samples with highest $p(\bs{x}$)}
  \end{subfigure}
  \hfill
  \begin{subfigure}{0.23\linewidth}
    \includegraphics[width=1\columnwidth]{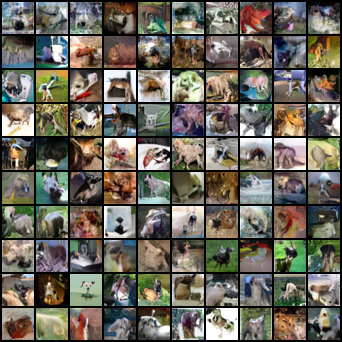}
    \caption{Samples with lowest $p(\bs{x}$)}
  \end{subfigure}
  \hfill
  \begin{subfigure}{0.23\linewidth}
    \includegraphics[width=1\columnwidth]{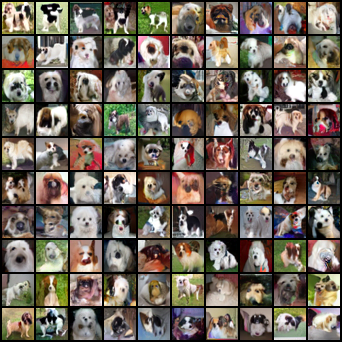}
    \caption{Samples with highest $p(y|\bs{x}$)}
  \end{subfigure}
  \hfill
  \begin{subfigure}{0.23\linewidth}
    \includegraphics[width=1\columnwidth]{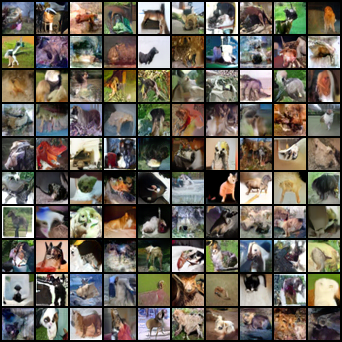}
    \caption{Samples with lowest $p(y|\bs{x}$)}
  \end{subfigure}
  \hfill
\caption{SADA-JEM generated class-conditional samples of \textbf{Dog}.}
\label{figure:sajem_app_class_5}
\end{figure*}

\begin{figure*}
  \centering
  \begin{subfigure}{0.23\linewidth}
    \includegraphics[width=1\columnwidth]{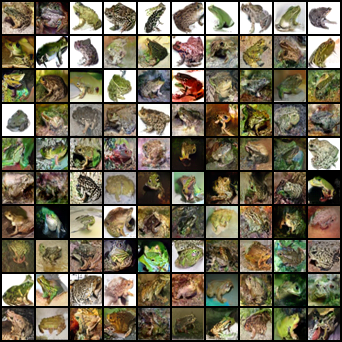}
    \caption{Samples with highest $p(\bs{x}$)}
  \end{subfigure}
  \hfill
  \begin{subfigure}{0.23\linewidth}
    \includegraphics[width=1\columnwidth]{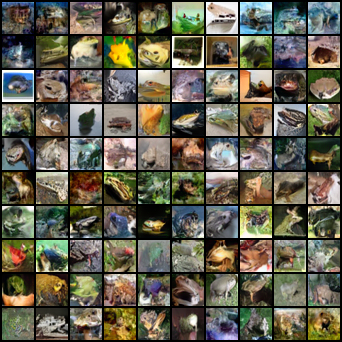}
    \caption{Samples with lowest $p(\bs{x}$)}
  \end{subfigure}
  \hfill
  \begin{subfigure}{0.23\linewidth}
    \includegraphics[width=1\columnwidth]{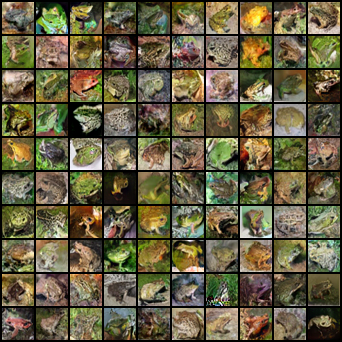}
    \caption{Samples with highest $p(y|\bs{x}$)}
  \end{subfigure}
  \hfill
  \begin{subfigure}{0.23\linewidth}
    \includegraphics[width=1\columnwidth]{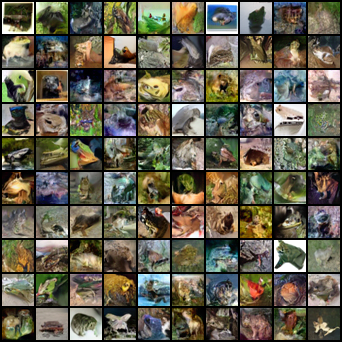}
    \caption{Samples with lowest $p(y|\bs{x}$)}
  \end{subfigure}
  \hfill
\caption{SADA-JEM generated class-conditional samples of \textbf{Frog}.}
\label{figure:sajem_app_class_6}
\end{figure*}

\begin{figure*}
  \centering
  \begin{subfigure}{0.23\linewidth}
    \includegraphics[width=1\columnwidth]{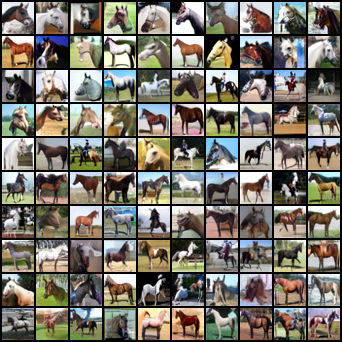}
    \caption{Samples with highest $p(\bs{x}$)}
  \end{subfigure}
  \hfill
  \begin{subfigure}{0.23\linewidth}
    \includegraphics[width=1\columnwidth]{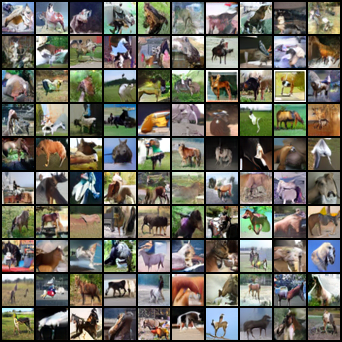}
    \caption{Samples with lowest $p(\bs{x}$)}
  \end{subfigure}
  \hfill
  \begin{subfigure}{0.23\linewidth}
    \includegraphics[width=1\columnwidth]{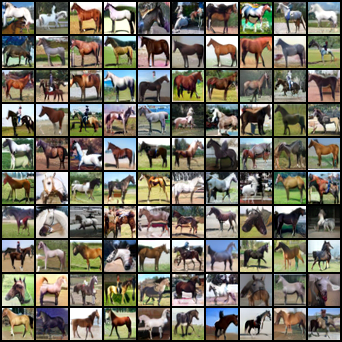}
    \caption{Samples with highest $p(y|\bs{x}$)}
  \end{subfigure}
  \hfill
  \begin{subfigure}{0.23\linewidth}
    \includegraphics[width=1\columnwidth]{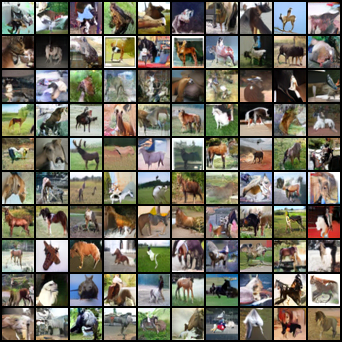}
    \caption{Samples with lowest $p(y|\bs{x}$)}
  \end{subfigure}
  \hfill
\caption{SADA-JEM generated class-conditional samples of \textbf{Horse}.}
\label{figure:sajem_app_class_7}
\end{figure*}

\begin{figure*}
  \centering
  \begin{subfigure}{0.23\linewidth}
    \includegraphics[width=1\columnwidth]{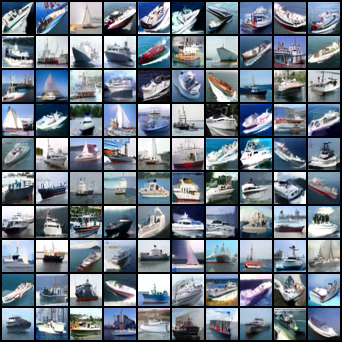}
    \caption{Samples with highest $p(\bs{x}$)}
  \end{subfigure}
  \hfill
  \begin{subfigure}{0.23\linewidth}
    \includegraphics[width=1\columnwidth]{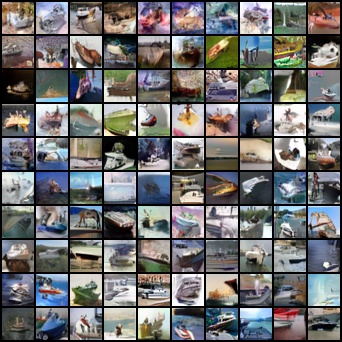}
    \caption{Samples with lowest $p(\bs{x}$)}
  \end{subfigure}
  \hfill
  \begin{subfigure}{0.23\linewidth}
    \includegraphics[width=1\columnwidth]{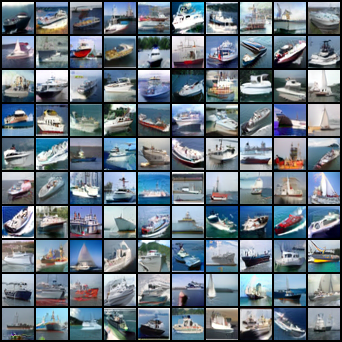}
    \caption{Samples with highest $p(y|\bs{x}$)}
  \end{subfigure}
  \hfill
  \begin{subfigure}{0.23\linewidth}
    \includegraphics[width=1\columnwidth]{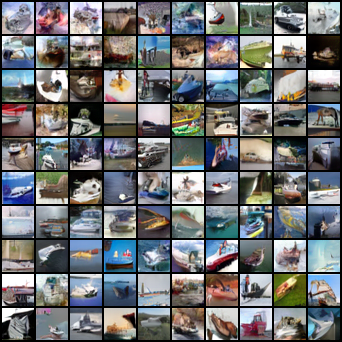}
    \caption{Samples with lowest $p(y|\bs{x}$)}
  \end{subfigure}
  \hfill
\caption{SADA-JEM generated class-conditional samples of \textbf{Ship}.}
\label{figure:sajem_app_class_8}
\end{figure*}

\begin{figure*}
  \centering
  \begin{subfigure}{0.23\linewidth}
    \includegraphics[width=1\columnwidth]{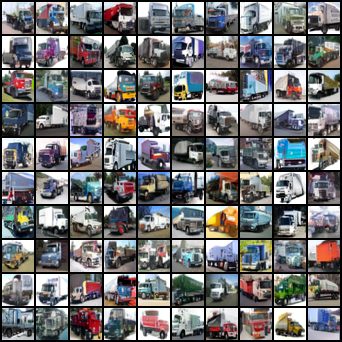}
    \caption{Samples with highest $p(\bs{x}$)}
  \end{subfigure}
  \hfill
  \begin{subfigure}{0.23\linewidth}
    \includegraphics[width=1\columnwidth]{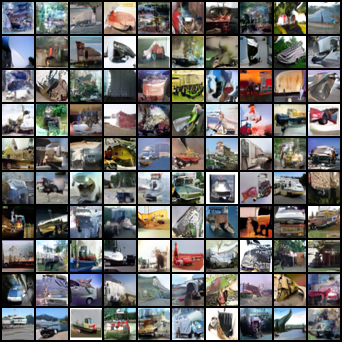}
    \caption{Samples with lowest $p(\bs{x}$)}
  \end{subfigure}
  \hfill
  \begin{subfigure}{0.23\linewidth}
    \includegraphics[width=1\columnwidth]{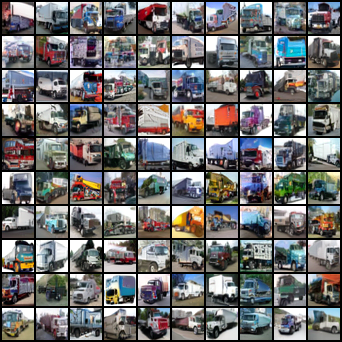}
    \caption{Samples with highest $p(y|\bs{x}$)}
  \end{subfigure}
  \hfill
  \begin{subfigure}{0.23\linewidth}
    \includegraphics[width=1\columnwidth]{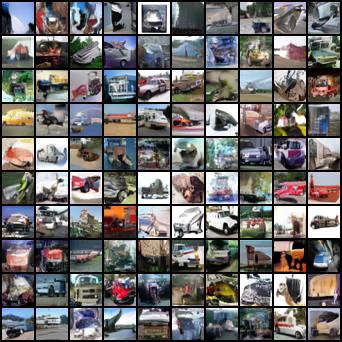}
    \caption{Samples with lowest $p(y|\bs{x}$)}
  \end{subfigure}
  \hfill
\caption{SADA-JEM generated class-conditional samples of \textbf{Truck}.}
\label{figure:sajem_app_class_9}
\end{figure*}

Additional SADA-JEM generated class-conditional (best and worst) samples of CIFAR10 are provided in Figures~\ref{figure:sajem_app_class_0}-\ref{figure:sajem_app_class_9}. 


\end{document}